\def\year{2021}\relax
\documentclass[letterpaper]{article} % DO NOT CHANGE THIS
\usepackage{aaai21}  % DO NOT CHANGE THIS
\usepackage{times}  % DO NOT CHANGE THIS
\usepackage{helvet} % DO NOT CHANGE THIS
\usepackage{courier}  % DO NOT CHANGE THIS
\usepackage[hyphens]{url}  % DO NOT CHANGE THIS
\usepackage{graphicx} % DO NOT CHANGE THIS
\usepackage{natbib}  % DO NOT CHANGE THIS AND DO NOT ADD ANY OPTIONS TO IT
\usepackage{caption} % DO NOT CHANGE THIS AND DO NOT ADD ANY OPTIONS TO IT
\title{Towards Understanding the Influence of Individual Clients in Federated Learning}
\author {
        Yihao Xue \textsuperscript{\rm 1},
        Chaoyue Niu \textsuperscript{\rm 1},
        Zhenzhe Zheng \textsuperscript{\rm 1},\\
        Shaojie Tang \textsuperscript{\rm 2},
        Chengfei Lyu \textsuperscript{\rm 3},
        Fan Wu \textsuperscript{\rm 1},
        and Guihai Chen \textsuperscript{\rm 1}\\
}
\affiliations {
    \textsuperscript{\rm 1} Shanghai Jiao Tong University,\\
    \textsuperscript{\rm 2} The University of Texas at Dallas,\\
    \textsuperscript{\rm 3} Alibaba Group \\
    yh\_xue@outlook.com, rvince@sjtu.edu.cn, zhengzhenzhe@sjtu.edu.cn,\\
    shaojie.tang@utdallas.edu, chengfei.lcf@alibaba-inc.com, fwu@cs.sjtu.edu.cn, gchen@cs.sjtu.edu.cn\\
}

\usepackage{graphicx}

\usepackage{algorithm}
\usepackage{algorithmicx}
\usepackage{algpseudocode}
\usepackage{amsmath}
\usepackage{booktabs}
\usepackage{multicol}
\usepackage{subfigure}
\usepackage{graphicx}
\usepackage{textcomp}
\usepackage{xcolor}
\usepackage{amsmath,amssymb,amsfonts}
\usepackage{algpseudocode}
\usepackage{courier}
\usepackage{appendix}
\usepackage{amsthm}

\newtheorem{assumption}{Assumption}
\newtheorem{theorem}{Theorem}

\newtheorem{definition}{Definition}

\algdef{SE}[VARIABLES]{Variables}{EndVariables}
   {\algorithmicvariables}
   {\algorithmicend\ \algorithmicvariables}
\algnewcommand{\algorithmicvariables}{\textbf{static variables}}

\usepackage{xpatch}
\xpatchcmd{\algorithmicx}% <cmd>
  {\newcommand{\State}{\ALC@it}}% <search>
  {\newcommand{\State}{\@ifstar\Statestar\Statenostar}}% <replace>
  {}{}% <success><failure>
\newcommand{\Statestar}{\item[]}
\newcommand{\Statenostar}{\ALC@it}

\makeatother
\begin{document}
\maketitle
\begin{abstract}
Federated learning allows mobile clients to jointly train a global model without sending their private data to a central server. Extensive works have studied the performance guarantee of the global model, however, it is still unclear how each individual client influences the collaborative training process. In this work, we defined a new notion, called {\em Fed-Influence}, to quantify this influence over the model parameters, and proposed an effective and efficient algorithm to estimate this metric. In particular, our design satisfies several desirable properties: (1) it requires neither retraining nor retracing, adding only linear computational overhead to clients and the server; (2) it strictly maintains the tenets of federated learning, without revealing any client's local private data; and (3) it works well on both convex and non-convex loss functions, and does not require the final model to be optimal. Empirical results on a synthetic dataset and the FEMNIST dataset demonstrate that our estimation method can approximate Fed-Influence with small bias. Further, we show an application of Fed-Influence in model debugging.
\end{abstract}

\section{Introduction}
Federated learning ingeniously leverages a large amount of valuable data in a distributed manner, while mitigating systemic privacy risks \cite{DBLP:conf/aistats/McMahanMRHA17, kairouz2019advances}. In such a setting, the training data are stored on multiple decentralized clients, who share a common global model and train it collaboratively. There is a central server that orchestrates the whole training process. In each round, the server collects local models from some eligible clients and use them to update the global model.

In this paper, we consider a significantly important but less considered problem in federated learning: {\em how does each client influence the global model?} Finding the answer to this question is meaningful in several aspects. On one hand, it helps us understand where the model comes from by explaining the prediction of the current global model in terms of clients. On the other hand, it provides quantitative insights into the roles of individual clients in federated learning, and informs us whether the existence of a certain client benefits the global model. This is critical to a fair credit/reward allocation, and, more importantly, debugging for federated learning. 
%In a centralized setting, modelers can directly check the data when the model is misbehaving, while in federated learning they suffer from a lack of data inspection. 
Thus, a fine-grained understanding of clients' influence further facilitates the exclusion of negative-influence clients or dynamically requires them to check their local data, thereby improving model performance. All of these are important to the explanation and robustness of federated learning, and also help sustain long-term user participation.

There already exists a classical statistics notion of ``influence" in centralized learning, which evaluates the effect that the absence of an individual sample has on a model. To measure this influence, one should conduct a leave-one-out test \cite{cook1977detection}:  retrain the model over the training set with one certain sample removed, and compare this model with that trained on the full dataset. A notion from robust statistics, called ``influence function" \cite{jaeckel1972infinitesimal,hampel1974influence,8666943c14594ea88ca7f492887941c7}, was introduced to avoid retraining by measuring the change in the model caused by slightly changing the weight of one sample and using  quadratic approximation combined with a Newton step \cite{cook1982residuals}. \citet{DBLP:conf/icml/KohL17} leveraged influence function in modern machine learning settings and developed an efficient and simple implementation using second-order optimization techniques. \citet{NIPS2019_8674} went beyond convexity and optimality, which are two important assumptions in \cite{DBLP:conf/icml/KohL17}. \citet{NIPS2019_8767} studied the effects of removing a group of data points, which is analogous to removing a client that holds a subset of training data in federated learning, except that their study is still in a centralized setting. \citet{khanna2019interpreting} applied Fisher kernels along with sequential Bayesian quadrature to identify a subset of training examples that are most responsible for a given set of predictions and recovered \cite{DBLP:conf/icml/KohL17} as a special case.

The existing works above focused on centralized learning, and we are the first to consider a similar problem, the influence of individual clients, under a brand new framework, namely federated learning. There are some essential differences between centralized learning and federated learning which raise several design challenges: (1) The server in centralized learning, as the influence evaluator, has the full control over the considered sampling data, while in federated learning, the server would not be able to access clients' raw data because of the privacy requirement; (2) the clients in federated learning may not always be available, due to the unreliable network connection. This implies that the server cannot communicate with a certain client at any desired time; and (3) the computing resources of mobile clients are limited, for which reason we should not bring too many additional computational burdens to clients when measuring their influence.

Due to the first difference, sample-level influence in centralized learning cannot be applied to measuring the influence of individual clients in federated learning. In this work, we turn to client-level influence measurement by investigating the effect of removing a client. In addition, works in centralized learning mainly focus on the influence on a model's testing loss, while we consider the influence on the parameter of the global model for the following two reasons: (1) In centralized learning, to cut down the time complexity, some techniques can be applied to obtain influence on loss without computing that on parameter \cite{NIPS2019_8674}. However, in federated learning, computing influence on parameter is unavoidable because of the second and third differences mentioned earlier. Please refer to the supplement \footnote{The supplement is available from https://drive.google.com/file\\/d/1zFefHCAYiv5DPJ6nDVjgLUeO4yOoAFlV/view?usp=sharing. } for detailed reasoning; and (2) influence on parameter is more fundamental and powerful than that on loss. With the knowledge of influence on parameter, we can easily derive the influence of individual clients in terms of various metrics for evaluating models, such as loss, accuracy, precision, etc.

\subsubsection{Our contributions.} (1) To the best of our knowledge, we are the first to consider client-level influence in federated learning; (2) we propose a basic estimator for individual clients' influence on model parameter by leveraging the relationship between the global models in two consecutive communication rounds. Guided by the error analysis, we extend the basic design to support both convex and non-convex loss functions. We also develop an efficient implementation, bringing only slight communication and computation overhead to the server and the clients (in fact no extra computation overhead for the client); and (3) empirical studies on a synthetic dataset and the FEMNIST dataset~\cite{DBLP:journals/corr/abs-1812-01097} demonstrate the effectiveness of our method. The estimation error observed in experiments indicates both the necessity and accuracy of our method. Based on the influence on parameter, we further derive the influence on model performance and observe that it was well estimated. In particular, the Pearson correlation coefficient between the estimated influence on loss and the ground truth achieves 0.62 in the most difficult setting. We also leverage influence on model performance for client valuation and client cleansing. %which achieves a 0.53\% increase in accuracy by removing 5\% of the clients in the most difficult setting.

%\begin{figure}[!t]
%\centering
%\includegraphics[width=0.95\columnwidth]{images//flowchart.pdf}
%\caption{An overview of our work.}
%\label{fig:flowchart}
%\end{figure}
%
%Fig.~\ref{fig:flowchart} gives an overview of the paper structure, which is composed of three parts. The black part depicts the original framework of federated learning introduced in Subsection~\ref{sec:fedml}. The blue part shows our design in Sections~\ref{sec:estimator} and \ref{sec:modification} to measure client-level influence, which works parallel with the model training. The orange part illustrates how influence measurement can benefit model performance, and the detailed results are presented in Section~\ref{sec:extension}.

\section{Problem Formulation}
\subsection{Federated Learning}\label{sec:fedml}
We first introduce some necessary notations. Let $\mathcal{C}$ denote the set of all the clients, and  $\mathcal{D}^k$ denote the local dataset of client $k \in \mathcal{C}$ with $n_k$ samples. The set $\mathcal{D}=\bigcup_{k\in \mathcal{C}} \mathcal{D}^k$ is the full training set. For any set of clients $\mathcal{C}'$, we use  $N(\mathcal{C}')=\sum_{k\in \mathcal{C}'}n_k$ to denote the total size of these clients' datasets. $\mathcal{L}({\bf w}, z)$ denotes the loss function over a model ${\bf w}$ and a sample $z$. In addition, $\mathcal{L}({\bf w}, \mathcal{D}^k)=\frac{1}{n_k}\sum_{z\in \mathcal{D}^k}\mathcal{L}({\bf w}, z)$ denotes the empirical loss over a model ${\bf w}$ and a dataset $\mathcal{D}^k$. Then, we consider the following optimization task of federated learning:
\begin{align}\label{eq:task}
\min_{\bf w\in \mathbb{R}^p} \left\{\mathcal{L}({\bf w}, \mathcal{D})=\sum_{k\in\mathcal{C}}\frac{n_k}{N(\mathcal{C})}\mathcal{L}({\bf w}, \mathcal{D}^k)\right\},
\end{align}
where the global loss function $\mathcal{L}({\bf w}, \mathcal{D})$ is the weighted average of the local functions $\mathcal{L}({\bf w}, \mathcal{D}^k)$ with the weight proportioning to the size of each client's local dataset. In this work, we consider a standard algorithm, federated averaging (FedAvg) \cite{DBLP:conf/aistats/McMahanMRHA17}, to solve the optimization problem in (\ref{eq:task}). Although there are some other variants, such as FedBoost \cite{49366}, FedNova \cite{DBLP:journals/corr/abs-2007-07481}, FetchSGD \cite{DBLP:journals/corr/abs-2007-07682}, FedProx \cite{DBLP:conf/mlsys/LiSZSTS20}, and SCAFFOLD \cite{DBLP:journals/corr/abs-1910-06378}, FedAvg is the first and the most widely used one. As a result, we see FedAvg as our basic block, which executes as follows. In the initial stage, the server randomly initializes a global model ${\bf w}_0$. Then, the training process is orchestrated by repeating the following two steps within each communication round $t$ from $1$ to $T$:
\begin{itemize}
\item {\bf Local training.} The server selects a random set of clients $\mathcal{C}_t$ as participants in this round. Each participant $k\in \mathcal{C}_t$ then downloads from the server the latest global model ${\bf w}_{t-1}$, \emph{i.e.}, the output model from the previous round $t-1$. Then, the client $k$ performs local updates for each local iteration $i$ from $1$ to $m$:
\begin{equation}
\label{eq:local_update}
{\bf w}^k_{t, i}\leftarrow {\bf w}^k_{t, i-1}-\eta \nabla_w \mathcal{L}\left({\bf w}^k_{t, i-1}, \mathcal{D}^k\right),
\end{equation}
with the starting local model ${\bf w}_{t, 0}^k$ initialized as ${\bf w}_{t-1}$. In addition, $\eta$ denotes the learning rate.
%, and $m$ denotes the number of local iterations.
\item {\bf Model aggregation.} Participants in round $t$ upload their updated local models. The server aggregates (takes a weighted average of) the local models to a new global model ${\bf w}_t$:
\begin{equation}
\label{eq:aggregation}
{\bf w}_t\leftarrow \sum_{k\in \mathcal{C}_t}\frac{n_k}{N(\mathcal{C}_t)}{\bf w}^k_{t,m},
\end{equation}
where the weight of client $k$ is the size of her dataset $n_k$.
\end{itemize}

\subsection{Fed-Influence}
To express the client-level influence on federated learning clearly, we introduce a new notation ${\bf w}_{t}\left(\mathcal{C}' \rightarrow\mathcal{C}'\backslash \{c\}\right)$, which represents the aggregated model in round $t$ with a client set $\mathcal{C}'$ replaced with $\mathcal{C}'\backslash \{c\}$, where $\mathcal{C}'\in \{\mathcal{C}_1, \mathcal{C}_2, \mathcal{C}_3, \dots, \mathcal{C}_T, \mathcal{C}\}$. For example, ${\bf w}_{10}\left(\mathcal{C}_5 \rightarrow\mathcal{C}_5\backslash \{c\}\right)$ represents the resulting model we get at the end of round $10$ if we remove the client $c$ in round $5$. 
One special case is that  $\mathcal{C}' = \mathcal{C}_t$, \emph{i.e.}, 
\begin{equation}
\label{eq:aggregation_loo}
{\bf w}_{t}\left(\mathcal{C}_{t} \rightarrow\mathcal{C}_t\backslash \{c\}\right) = \sum_{k \in \mathcal{C}_t \backslash \{c\}}\frac{n_k }{N\left(\mathcal{C}_t \backslash \{c\}\right)} {\bf w}_{t, m}^k.
\end{equation}
Another special case is ${\bf w}_{t}\left(\mathcal{C}\rightarrow\mathcal{C}\backslash \{c\}\right)$, where we permanently remove the client $c$, \emph{i.e.}, replace $ \mathcal{C}_{t}$ with $\mathcal{C}_{t}\backslash\{c\}$ for all the rounds $t\in [T]$.

%To quantify the influence of individual clients,
With these notations, we next give the following definition of {\em Fed-Influence on Parameter (FIP)}.
\begin{definition}
We refer to the change in  model parameters due to removing a client $c$ from $\mathcal{C}$ as Fed-Influence on Parameter (FIP) of client $c$, denoted by $\epsilon_t^{-c,*}$:
\begin{equation}
\epsilon_t^{-c,*}\overset{\rm def}{=}{\bf w}_t\left(\mathcal{C} \rightarrow\mathcal{C} \backslash \{c\} \right)-{\bf w}_t.
\end{equation}
\end{definition}

Based on FIP, we can extend the notion to measure the influence on model performance, such as the metrics of accuracy, cross-entropy loss, precision, recall, mean squared error (MSE), and etc. We can derive the influence on any of these metrics from FIP. Suppose  $\mathcal{F}$ is the loss function for a certain metric over a test set $\mathcal{D}_{test}$, we can compute
\begin{align}
\mathcal{F}\left({\bf w}_t + \epsilon_t^{-c,*}, \mathcal{D}_{test}\right)-\mathcal{F}\left({\bf w}_t, \mathcal{D}_{test}\right)
\end{align}
as the Fed-Influence over the metric. In Sections \ref{sec:basic_exp} and \ref{sec:extension}, we will focus on two most widely used metrics, loss and accuracy, which corresponds to Fed-Influence on Loss (FIL) and Fed-Influence on Accuracy (FIA), respectively.

The exact value of FIP for a client can only be obtained by conducting leave-one-out test: retrain the model by removing the client, and compare the retrained model with the model trained on the full client set. However, it is prohibitively inefficient to rerun the whole federated learning process, especially when we intend to measure the influence of each client, implying the number of rerunning being the total number of clients.

\section{Basic Estimator}\label{sec:estimator}
We now derive an estimator for FIP  $\epsilon_t^{-c,*}$ to avoid retraining. We start by rewriting the expression of $\epsilon_t^{-c,*}$ as follows
\begin{align}
\label{eq:decomposition}
\nonumber
\epsilon_t^{-c,*} & \overset{\rm def}{=}~{\bf w}_{t}\left(\mathcal{C} \rightarrow\mathcal{C}\backslash \{c\}\right) - {\bf w}_t\\
\nonumber
&={\bf w}_{t}\left(\mathcal{C} \rightarrow\mathcal{C}\backslash \{c\}\right) - {\bf w}_{t}\left(\mathcal{C}_{t} \rightarrow\mathcal{C}_t\backslash \{c\}\right)\\
\nonumber
&+{\bf w}_{t}\left(\mathcal{C}_{t} \rightarrow\mathcal{C}_t\backslash \{c\}\right)- {\bf w}_t\\
\nonumber
&=\underbrace{\sum_{k \in \mathcal{C}_t \backslash \{c\}}\frac{n_k}{N(\mathcal{C}_t \backslash \{c\})}~\underbrace{\left({\bf w}_{t, m}^k\left(\mathcal{C} \rightarrow\mathcal{C}\backslash \{c\}\right) - {\bf w}_{t, m}^k \right)}_{\text{local sequential influence}}}_{\text{sequential influence}} \\
&+\underbrace{{\bf w}_{t}\left(\mathcal{C}_{t} \rightarrow\mathcal{C}_t\backslash \{c\}\right)- {\bf w}_t}_{\text{combinatorial influence}}
\end{align}
Equation~(\ref{eq:decomposition}) shows that $\epsilon_t^{-c,*}$ comprises three parts: (1) {\em Local sequential influence}: the influence that removing client $c$ from the client set $\mathcal{C}$ on the local model of any other participating client $k\in \mathcal{C}_t \backslash \{c\}$ in round $t$. We regard it as ``sequential" influence because it can be derived from $\epsilon_{t-1}^{-c,*}$, the influence in the previous round; (2) {\em Sequential influence}: the weighted average of local sequential influence; and (3) {\em Combinatorial influence}: the influence of removing $c$ merely from $\mathcal{C}_t$ in the round $t$. The combinatorial influence is ``combinatorial" as it is independent of $\epsilon_{t-1}^{-c,*}$.

We next dissect how to compute local sequential influence and combinatorial influence. The combinatorial one can be easily obtained using (\ref{eq:aggregation}) and (\ref{eq:aggregation_loo}). The local sequential influence is more complicated. Given that the reasons behind the difference between ${\bf w}_{t, m}^k\left(\mathcal{C} \rightarrow\mathcal{C}\backslash \{c\}\right)$ and ${\bf w}_{t, m}^k$ is that they are locally updated from different initial models, ${\bf w}_{t-1}\left(\mathcal{C} \rightarrow\mathcal{C}\backslash \{c\}\right)$ and ${\bf w}_{t-1}$, respectively. 
We take a look at the first local iteration at round $t$.
we estimate the term by applying first-order Taylor approximation and the chain rule:
\begin{align}
\label{eq:chain_rule}
\nonumber
&{\bf w}_{t, m}^k\left(\mathcal{C} \rightarrow\mathcal{C}\backslash \{c\}\right) - {\bf w}_{t, m}^k\\
\approx~& \frac{\partial {\bf w}_{t,m}^k}{\partial {\bf w}_{t,m-1}^k}\frac{\partial {\bf w}_{t,m-1}^k}{\partial {\bf w}_{t,m-2}^k}\dots \frac{\partial {\bf w}_{t,1}^k}{\partial {\bf w}_{t,0}^k}\Delta {\bf w}_{t,0}^k
\end{align}
where $\Delta {\bf w}_{t,0}^k={\bf w}_{t,0}^k\left(\mathcal{C} \rightarrow\mathcal{C}\backslash \{c\}\right) - {\bf w}_{t,0}^k=\epsilon_{t-1}^{-c,*}$. According to the update rule in Equation~\ref{eq:local_update}, with the assumption that $\mathcal{L}({\bf w}, \mathcal{D}^k)$ is twice differentiable, we obtain
\begin{align}
\label{eq:second_derivative}
\frac{\partial {\bf w}_{t,i}^k}{\partial {\bf w}_{t,i-1}^k}=\mathbf{I}-\eta \mathbf{H}_{t,i-1}^k,
\end{align}
where $\mathbf{H}_{t,i}^k \overset{\rm def}{=} \nabla^2_w \mathcal{L}({\bf w}_{t,i}^k,\mathcal{D}^k)$. By combining Equations~\ref{eq:decomposition}, \ref{eq:chain_rule} and \ref{eq:second_derivative}, we get an estimator of $\epsilon_t^{-c,*}$, denoted by $\epsilon_t^{-c}$
\begin{align}
\label{eq:estimator}
\epsilon_t^{-c} \overset{\rm def}{=} \mathbf{M}_t^{-c}\epsilon_{t-1}^{-c}+{\bf w}_{t}\left(\mathcal{C}_{t} \rightarrow\mathcal{C}_t\backslash \{c\}\right)- {\bf w}_t,
\end{align}
where
\begin{equation}
\label{eq:M_original}
\mathbf{M}_t^{-c}\overset{\rm def}{=}\sum_{k\in\mathcal{C}_t \backslash\{c\}}\frac{n_k}{N(\mathcal{C}_t \backslash\{c\})}\prod_{i=0}^{m-1}(\mathbf{I}-\eta \mathbf{H}_{t, i}^k).
\end{equation}
By recalling that the initial model ${\bf w}_0$ is randomly initialized by the server, we have $\epsilon_0^{-c}=\mathbf{0}$. Then the estimator $\epsilon_t^{-c}$ can be computed iteratively using Equation~\ref{eq:estimator}.

We finally take a close look at the relation between the estimation error and the iteration $t$. We give a uniform bound on the error in both convex and non-convex cases under the following assumptions.

\begin{assumption} \label{assump:bounded_hessian}
There exists $\lambda$ and $\Lambda$ such that $\lambda I \preccurlyeq\nabla^2\mathcal{L} \preccurlyeq\Lambda I$.
\end{assumption}
\begin{assumption} \label{assump:bounded_change}
The norm of $\epsilon_t^{-c,*}$ is bounded by $C$, for $t\in [T]$ and $c\in\mathcal{C}$.
\end{assumption}
Note that in Assumption~\ref{assump:bounded_hessian}, there is no constraint on the values of $\lambda$ and $\Lambda$ except $\lambda\leq\Lambda$, which means the loss function is not necessarily convex. Next we give Theorem~\ref{theo:error}, the proof of which is provided in the supplement.

\begin{theorem}
\label{theo:error}
With Assumptions~\ref{assump:bounded_hessian} and \ref{assump:bounded_change}, the error of the estimator is bounded by
\begin{equation}
\label{eq:error_bounding}
\delta_t^{-c}\overset{\rm def}{=}\|\epsilon_t^{-c,*}-\epsilon_t^{-c}\|\leq\frac{1-\gamma^t}{1-\gamma}o(C), \quad \forall t>0,
\end{equation}
where $o$ is the little-o notation, and $\gamma=\alpha^m, \alpha=\max\{|1-\eta \lambda|, |1-\eta \Lambda|\}$.
\end{theorem}
From (\ref{eq:error_bounding}), we can observe that the bound is in the format of the sum of geometric series. An intuitive explanation is that each time we use (\ref{eq:estimator}), the error in the previous round is scaled by $\mathbf{M}_t^{-c}$, and added a newly introduced error in this round, similar to summing a geometric series. Finally, there are three different cases depending on the relation between $\gamma$ and $1$:
\begin{itemize}
\item {\bf Case 1} ($\gamma < 1$): In this case, $\lambda>0$ and $0<\eta<\frac{2}{\Lambda}$, where the loss function is strongly-convex and the learning rate is small enough. This is the most ideal case, where the bound can be further scaled to $\frac{1}{1-\gamma}o(C)$, which is independent of $t$.

\item {\bf Case 2} ($\gamma=1$): In this case, either $\lambda=0$ and $\eta\leq\frac{2}{\Lambda}$ or $\lambda\geq0$ and $\eta=\frac{2}{\Lambda}$. The former situation is more common, with a convex loss function and an appropriate learning rate. Then the error is $o(C)t$, linear with $t$.

\item {\bf Case 3} ($\gamma>1$): In this case, $\lambda<0$ or $\eta>\frac{2}{\Lambda}$, where either the loss function is non-convex or the learning rate is too large. Then the error bound is exponential with $t$, making estimation ineffective.
\end{itemize}

\section{Improving Robustness and Efficiency}\label{sec:modification}
In this section, we improve the basic estimator from two aspects: one is to improve the robustness of the method in non-convex case, and the other is to cut down the high cost due to computing  Hessian matrix. 
%Algorithm sketches are deferred to the supplement.

\subsection{Layer-Wise Examination and Truncation}
The analysis in Section~\ref{sec:estimator} reveals that the basic estimator can have a large error when the loss function is non-convex. The non-convex case is quite common in federated learning for deep learning tasks \cite{yu2019parallel, haddadpour2019local}. We thus propose layer-wise examination and truncation (LWET for short).
%, the details of which and the intuitions are shown as follows.
\subsubsection{Truncation.}
Our primary goal is to avoid an exponential error in Case 3, which results from the estimation of sequential influence. We consider a counterpart, where the sequential influence is completely omitted, {\em i.e.}, $\epsilon_t^{-c}={\bf w}_{t}\left(\mathcal{C}_{t} \rightarrow\mathcal{C}_t\backslash \{c\}\right)- {\bf w}_t$ (we call it the {\em truncated estimator} for simplicity), and find that the error becomes independent of $t$, as shown in the following theorem.
\begin{theorem}
\label{theo:error_trunc}
With the sequential influence omitted, we get the bound of error as follows:
\begin{equation}
\nonumber
\delta_t^{-c}\leq\gamma C+o(C).
\end{equation}
\end{theorem}

\begin{figure}[!t]
\centering
\subfigure[]{
\begin{minipage}[t]{0.5\columnwidth}
\centering
\includegraphics[width=1.5in]{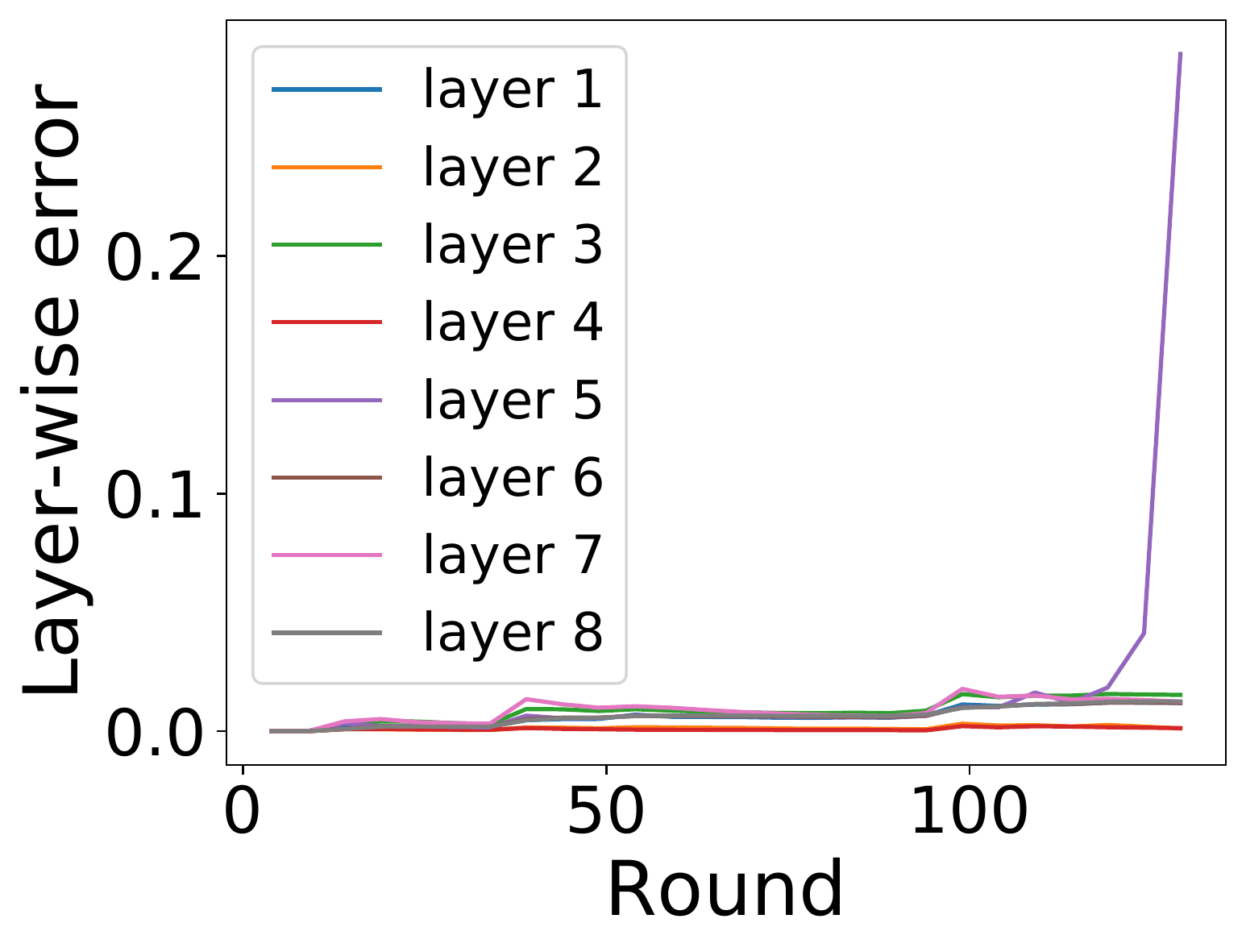}
%\caption{fig1}
\label{fig:overflow:error}
\end{minipage}%
}%
\subfigure[]{
\begin{minipage}[t]{0.5\columnwidth}
\centering
\includegraphics[width=1.5in]{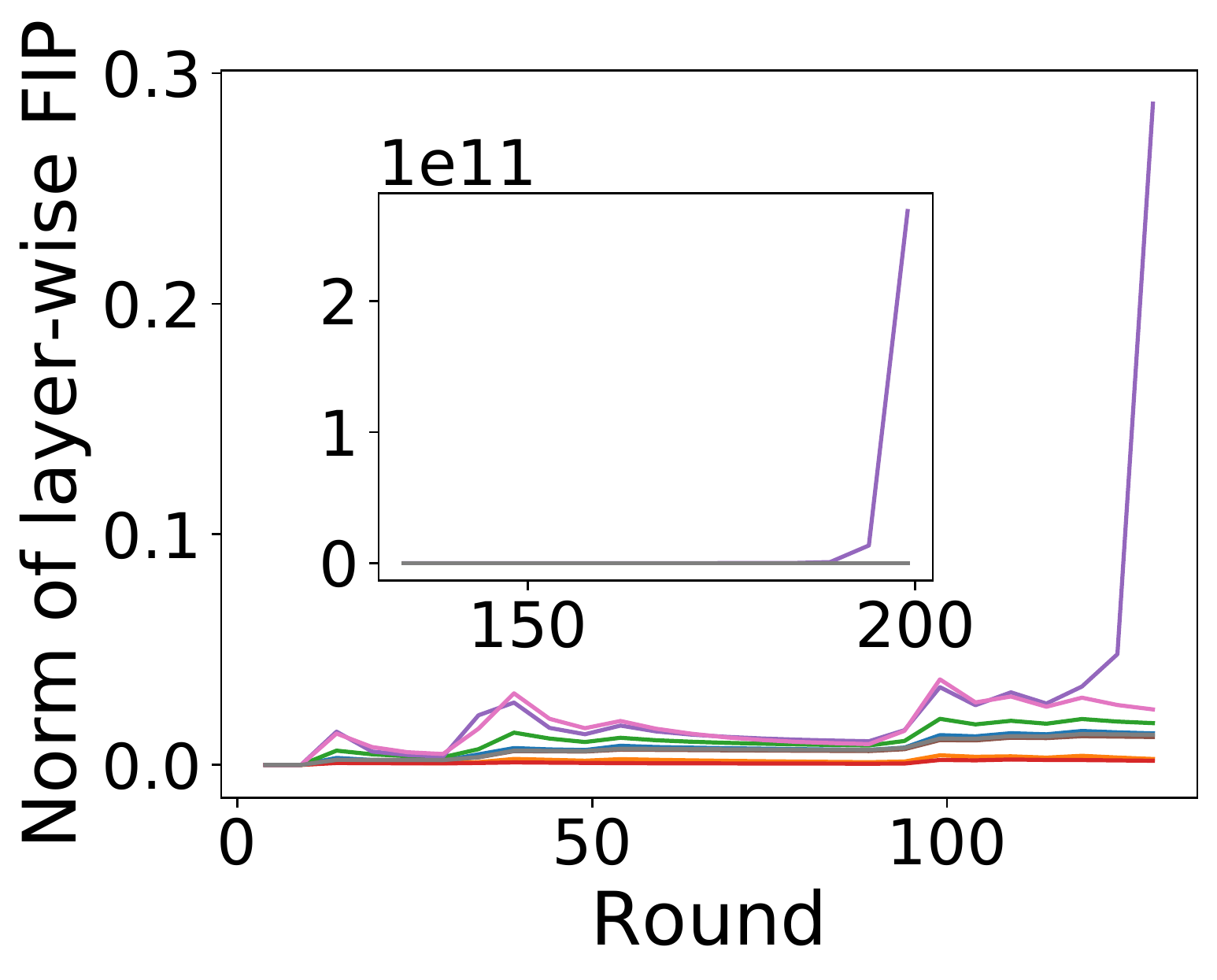}
%\caption{fig2}
\label{fig:overflow:norm}
\end{minipage}%
}%
\centering
\caption{The experimental result on a toy model (setting 2 in Table \ref{tab:settings}), a CNN without activation function. (a) The error and (b) the norm of estimated FIP both have a tendency to increase exponentially on layer 5, the first fully-connected layer.  Inset: the norm of estimated FIP from round 135 to 200.}
\label{fig:overflow}
\end{figure}

\subsubsection{Layer-wise Operation.} However, after truncation, a new problem arises: the truncated estimator will always be $\mathbf{0}$ at round $t$ as long as $c$ is not one of the participants. For example, in a setting where 5 clients are selected each round with 100 clients in total, there will be 95 clients with FIP being $\mathbf{0}$ each round, which does not make sense. To ensure accuracy while retaining as much information as possible, it is necessary to find a happy medium between the basic estimator and the truncated estimator, which instead partially omits the sequential influence. To achieve this, we first introduce layer-wise operation. We calculate only parts of the Hessian matrix, with the interaction between different layers ignored. We take a convolutional neural network (CNN) for example. Supposing that the parameter ${\bf w}$ is composed of ${\bf w}_{(j)}$, $j=1,2,\dots,8$, corresponding to conv-layer 1, bias of conv-layer 1, conv-layer 2, bias of conv-layer 2, dense-layer 1, bias of dense-layer 1, dense-layer 2, bias of dense-layer 2, respectively, {\em i.e.}, ${\bf w}=[{\bf w}_{(1)}^T, {\bf w}_{(2)}^T, {\bf w}_{(3)}^T, {\bf w}_{(4)}^T, {\bf w}_{(5)}^T, {\bf w}_{(6)}^T, {\bf w}_{(7)}^T, {\bf w}_{(8)}^T]^T$. For each layer $j$, we see $\mathcal{L}({\bf w})$ as function of ${\bf w}_{(j)}$ denoted by $\mathcal{L}_{(j)}({\bf w}_{(j)})$. Rather than computing the entire Hessian matrix $\mathbf{H}=\nabla^2_{w} \mathcal{L}({\bf w})$, we only calculate $\mathbf{H}_{(j)}=\nabla^2_{w_{(j)}} \mathcal{L}_{(j)}({\bf w}_{(j)})$ for each $j$, some smaller matrices on the diagonal of $\mathbf{H}$. Then the estimator in each layer $j$ is updated independently:
\begin{align}
%\label{eq:layer_wise_update}
\nonumber
\epsilon_{t,(j)}^{-c} \leftarrow \mathbf{M}_{t,(j)}^{-c}\epsilon_{t-1,(j)}^{-c}+{\bf w}_{t,(j)}\left(\mathcal{C}_{t} \rightarrow\mathcal{C}_t\backslash \{c\}\right)- {\bf w}_{t,(j)},
\end{align}
where $\mathbf{M}_{t,(j)}^{-c}=\sum_{k\in\mathcal{C}_t \backslash\{c\}}\frac{n_k}{N(\mathcal{C}_t \backslash\{c\})}\prod_{i=0}^{m-1}(\mathbf{I}-\eta \mathbf{H}_{t, i,(j)}^k)$. We can examine separately the property of the loss function $\mathcal{L}_{(j)}({\bf w}_{(j)})$ in each layer $j$, and use a layer-wise truncated estimator shown below only for layers in Case 3:
\begin{equation}
\label{eq:layer_wise_update_trunc}
\epsilon_{t,(j)}^{-c} \leftarrow {\bf w}_{t,(j)}\left(\mathcal{C}_{t} \rightarrow\mathcal{C}_t\backslash \{c\}\right)- {\bf w}_{t,(j)}.
\end{equation}
Then we put together layer-wise FIPs to get the complete FIP, {\em i.e.}, $\epsilon_{t}^{-c}=[(\epsilon_{t,(1)}^{-c})^T, (\epsilon_{t,(2)}^{-c})^T \dots]^T$.

Because the convexity and continuity of $\mathcal{L}_{(j)}({\bf w}_{(j)})$ vary among different layers, there are layers in Case 1 or Case 2 which still remain the sequential influence. Therefore the complete FIP is non-zero and contains much information even when $c\notin \mathcal{C}_t$. In an experiment conducted on a toy model (CNN 2 in Table~\ref{tab:settings}), we found the exponential error only exists in the first fully-connected layer, which is the only layer where we need to apply a layer-wise truncated estimator, as shown in Fig.~\ref{fig:overflow:error}.

\subsubsection{Examination.}
The next problem is how we can examine the property of loss function in each layer? We propose a mechanism which does not require any \emph{prior knowledge}: in particular, we compare $\|{\bf M}_{t,(j)}^{-c}\epsilon_{t-1,(j)}^{-c}\|$ with $\|\epsilon_{t-1,(j)}^{-c}\|$. If $\|{\bf M}_{t,(j)}^{-c}\epsilon_{t-1,(j)}^{-c}\|$ is larger at a certain round $r$, then, in all of the following rounds, {\em i.e.}, for all $t\geq r$, a layer-wise truncated estimator in Equation~\ref{eq:layer_wise_update_trunc} will be used in layer $j$.

This mechanism is aimed at examining a sufficient (but not necessary) condition for $\gamma>1$. Please refer to the supplement for the reason. In addition, this examination also avoids the overflow of the estimator itself. The upper bound of $\epsilon_{t-1}^{-c}$ is also exponential with $t$ in Case 3, which means there is always a risk for it to overflow as $t$ increases. This phenomenon can be observed in the experiment with the aforementioned toy model. As shown in Fig.~\ref{fig:overflow:norm}, the norm of estimated FIP on the first fully-connected layer increases sharply around the 125-th round, and then achieves an order of $10^{11}$ at round 200, indicating the failure of the algorithm.

Combining the three strategies, we get LWET. Because real-world applications rarely satisfy the Identically and Independently Distributed (IID) assumption and are likely to be non-IID in many ways \cite{zhao2018federated,DBLP:conf/mlsys/LiSZSTS20,DBLP:conf/iclr/LiHYWZ20, yan2020distributed}, where clients vary in the data distribution and the property of the local loss function, a more fine-grained version of LWET is recommended in these cases. We can examine the relationship between $\|\left(\prod_{i=0}^{m-1}(\mathbf{I}-\eta \mathbf{H}_{t, i,(j)}^k)\right)\epsilon^{-c}_{t,(j)}\|$ and $\|\epsilon^{-c}_{t,(j)}\|$. If the former is larger at one round, then we drop the local sequential influence on $k$ from that round on. Please refer to this algorithm in the supplement.

\subsection{Low-Cost Hessian Approximation}\label{sec:low-cost}
\subsubsection{Fisher information.}
Because the cross entropy loss is a negative log-likelihood, it is not difficult to obtain that $\mathop{\mathbb{E}}_{z\in \mathcal{D}^{'}}\left[\nabla_{w}\mathcal{L}({\bf w}^{*}, z) \nabla_{w}\mathcal{L}({\bf w}^{*}, z)^T\right]$, where ${\bf w}^{*}$ is the true parameter ({\em i.e.}, the model distribution under ${\bf w}^{*}$ equals to the underlying distribution), is at the form of Fisher information. And according to one of the alternative definition of Fisher information \cite{friedman2001elements,ly2017tutorial}, it can also be written as $\mathop{\mathbb{E}}_{z\in \mathcal{D}^{'}}[\nabla_{w}^2\mathcal{L}({\bf w}^{*}, z)]$. This means we can use $\nabla_{w}\mathcal{L}({\bf w}, z) \nabla_{w}\mathcal{L}({\bf w}, z)^T$ as an asymptotically unbiased estimation of $\nabla_{w}^2\mathcal{L}({\bf w}, z)$, since ${\bf w}$ gradually converges to ${\bf w}^{*}$ during the training.

Leveraging the fact that clients holds their own datasets, we let each client randomly select a given number, denoted by $N_s$, of gradients and use the empirical expectation of the outer product as an approximation to the Hessian, denoted by $\tilde{\mathbf{H}}_{t,i,(j)}^k$:
\begin{align}
\nonumber
\mathbf{H}_{t,i,(j)}^k\approx \tilde{\mathbf{H}}_{t,i,(j)}^k \overset{\rm def}{=} \frac{1}{N_s}\sum_{z\in \mathcal{S}_{t,i}^k}g({\bf w}_{t,i,(j)}^k, z)g({\bf w}_{t,i,(j)}^k, z)^T %\overset{\rm def}{=}\tilde{\mathbf{H}}_{t,i,(j)}^k,
\end{align}
where $\mathcal{S}_{t,i}^k$ is the set of $N_{s}$ samples randomly selected by client $k$ at the $i$-th local iteration in round $t$, and $g({\bf w}_{t,i,(j)}^k, z)=\nabla_{w_{(j)}}\mathcal{L}_{(j)}({\bf w}_{t,i,(j)}^k, z)$.

\subsubsection{Recursive computation.}
Simply introducing Fisher information cannot help cut down the cost, because the outer product is  $O(p^2)$, and the $O(p^2)$ matrix-vector or $O(p^3)$ matrix-matrix multiplication still exists, where $p$ is the size of the model. However, we can
actually circumvent these needless calculations. With $\mathbf{H}_{t,i,(j)}^k$ approximated by $\tilde{\mathbf{H}}_{t,i,(j)}^k$, the estimator of local sequential influence is
\begin{align}
\label{eq:closer_look_at_local_sequential_influence}
%\nonumber
\left(\prod_{i=0}^{m-1}\left(\mathbf{I}-\eta \frac{1}{N_s}\sum_{z\in \mathcal{S}_{t,i}^k}g({\bf w}_{t,i,(j)}^k, z) g({\bf w}_{t,i,(j)}^k, z)^T\right)\right) \epsilon_{t-1,(j)}^{-c}.
\end{align}
Instead of first calculating the cumprod on the left and then multiplying it by $\epsilon_{t-1,(j)}^{-c}$, the linear-cost method is based on a recursive computation: we initialize a vector $\sigma_{(j)}^k$ by $\sigma_{(j)}^k\leftarrow\epsilon_{t-1,(j)}^{-c}$, and then we update $\sigma_{(j)}^k$ using Equation~\ref{eq:update_sigma} repeatedly for $i$ from $0$ to $m-1$:
\begin{equation}
\label{eq:update_sigma}
\sigma_{(j)}^k\leftarrow \sigma_{(j)}^k-\frac{\eta}{N_s}\sum_{z\in \mathcal{S}_{t,i}^k}g({\bf w}_{t,i,(j)}^k, z) \bigg(g({\bf w}_{t,i,(j)}^k, z)^T \sigma_{(j)}^k\bigg).
\end{equation}
 $\sigma_{(j)}^k$ produced in the final iteration is the value of Equation~\ref{eq:closer_look_at_local_sequential_influence}. Note that this method requires the computation to take place on the server, because only the server has $\epsilon_{t-1,(j)}^{-c}$, and the clients need to do nothing other than randomly select and upload a certain number of local gradients. We show the efficiency of our method by comparing it with naive implementations in Table~\ref{tab:cost_comp}.

\begin{table}[!t]
\centering
\begin{tabular}{c| c c c}
\toprule[2pt]
 & Server & Client & Comm.\\
\midrule
Naive 1 & $K^2p^2$ & $m(np^2+p^3)$ & $p^2$\\
%\hline
Naive 2 & $K^2mp^2$ & $mnp^2$ & $mp^2$\\
%\hline
Our method & $K^2mN_{s}p$ & $0$ & $mN_{s}p$\\
\bottomrule[2pt]
\end{tabular}
\caption{Extra time complexity for the server, extra time complexity for each client, and extra communication complexity for each client. $n$ is the size of the local dataset and $K=|\mathcal{C}|$. In a naive implementation, operations where the Hessian matrix is involved, can be taken either on the clients or the server, corresponding to naive implementations 1 and 2, respectively. For naive implementation 1, each client $k$ has to compute $\prod_{i=0}^{m-1}(\mathbf{I}-\eta \mathbf{H}_{t, i}^k)$ locally, which requires matrix-matrix multiplications, and upload the result to the server. For naive implementation 2, each client $k$ uploads $\mathbf{H}_{t, i}^k$ for all $i$, and then the server computes $\mathbf{M}_t^{-c}\epsilon_{t-1}^{-c}$, which can be implemented with only matrix-vector multiplications.}
\label{tab:cost_comp}
\end{table}

\section{Fundamental Experiments}\label{sec:basic_exp}
\begin{table*}[!t]
\centering
\label{tab:settings}
\begin{tabular}{c| c c c c c c c c c}
\toprule[2pt]
 & Model & Dataset & Distribution & $\eta$ & $|\mathcal{C}|$ & $|\mathcal{C}_t|$ & $m$ & $T$ & $N_s$\\
\midrule
Setting 1 & LogReg & Synthetic & Non-IID, Unbalanced & 0.003 & 1000 & 10 & 5 & 1000 & 50\\
Setting 2 & CNN 1 & FEMNIST & IID, Balanced & 0.03 & 50 & 5 & 2 & 500 & 50\\
Setting 3 & CNN 2 & FEMNIST & Non-IID, Unbalanced & 0.02 & 100 & 10 & 2 & 2000 & 50\\
\bottomrule[2pt]
\end{tabular}
\centering
\caption{Detailed configuration of the three different settings. The two datasets are described in \citet{DBLP:journals/corr/abs-1812-01097}. The logistic regression model is the original one in ``Leaf". We made a little adjustment to the original CNN to create models with certain properties and scales; see details in the supplement.}
\label{tab:settings}
\end{table*}
In this section, we demonstrate two properties of our method: (1) LWET plays a vital role; and (2) Hessian approximation causes only a slight drop in the accuracy. Experiments are conducted on 64bit Ubuntu 18.04 LTS with four Intel i9-9900K CPUs and two NVIDIA RTX-2080TI GPUs, 200GB storage. We take ``Leaf" \cite{DBLP:journals/corr/abs-1812-01097}, a benchmarking framework for federated learning based on tensorflow. We evaluated our method on three settings, as shown in Table~\ref{tab:settings}. We used the softmax function at the output layer and adopted the cross entropy as the loss function. In setting 1, the loss function is convex but not strongly convex, and therefore it is in Case 2 ($\gamma=1$). In setting 2, although the toy model has no activation function, which makes it equivalent to a single-layer perceptron with convex loss function, results show that it is still in Case 3 ($\gamma>1$) because the learning rate is too large. And in setting 3, the loss function is non-convex and is therefore in Case 3, too.

\subsection{Influence on Parameter}
We use four different methods to obtain $\epsilon_t^{-c}$: (1) the basic estimator, (2) the estimator with only LWET, (3) the estimator with only Hessian approximation, and (4) the estimator with both LWET and Hessian approximation. We do not demonstrate results of all methods in each setting. In setting 1, we had tested all of the four methods, but found that the result produced with LWET is completely the same with that produced without LWET, which further validates that setting 1 is in Case 2. Therefore we only show two different results, the result based on exact Hessian and that based on approximated Hessian. In setting 2, we do not demonstrate the method with only Hessian approximation, because the basic estimator has already caused a terrible error, and the introduction of Hessian approximation will, no doubt, create an even larger error. In setting 3, we can only test the two methods that contain a Hessian approximation because the large model size and the limited resources on our device do not allow us to compute the exact Hessian.

\begin{figure}[!t]
\centering
\includegraphics[width=0.65\columnwidth]{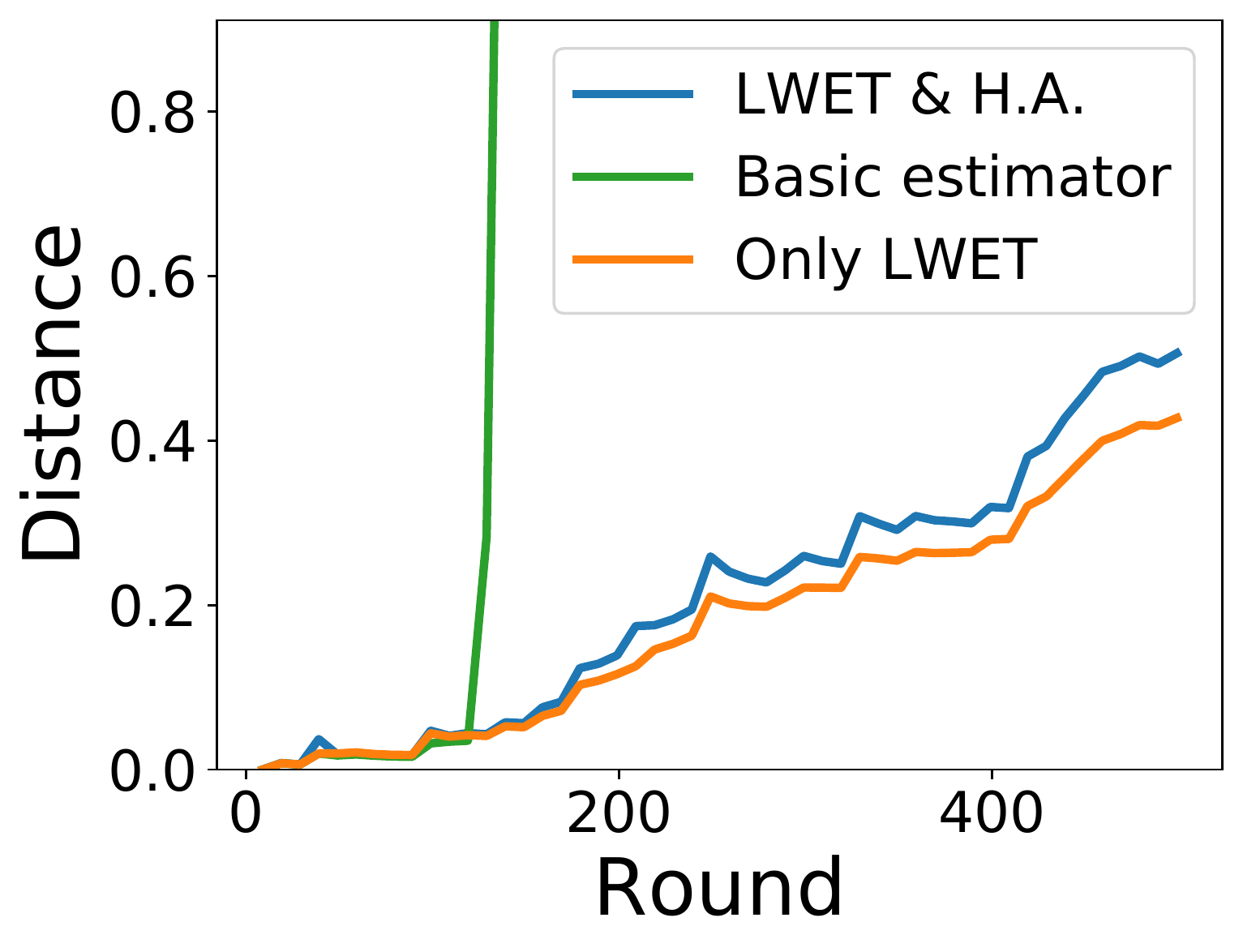}
\centering
\caption{The distance between exact and estimated FIP in setting 2.}
\label{fig:distance}
\end{figure}

We examine the error of the proposed method $\|\epsilon_t^{-c,*}-\epsilon_t^{-c}\|$, which is the distance between exact FIP and estimated FIP under $L_2$ norm. The exact FIP is obtained by conducting leave-one-out tests. We track the error of one randomly selected client's FIP and show the result in setting 2 in figure \ref{fig:distance}; results in the other two settings are provided as supplementary materials. Here we can see the necessity of LWET: without LWET there will be a sharp increase in the error at about the 120-th round (the error is caused by the first fully-connected layer as mentioned earlier in Fig.~\ref{fig:overflow}). Further taking a Hessian approximation only causes a slight increase in the error.

\subsection{Influence on Loss}\label{sec:fil}
\begin{figure*}[!t]
\centering
\subfigure[]{
\label{fig:fil:logreg_pearson}
\includegraphics[width=0.568\columnwidth]{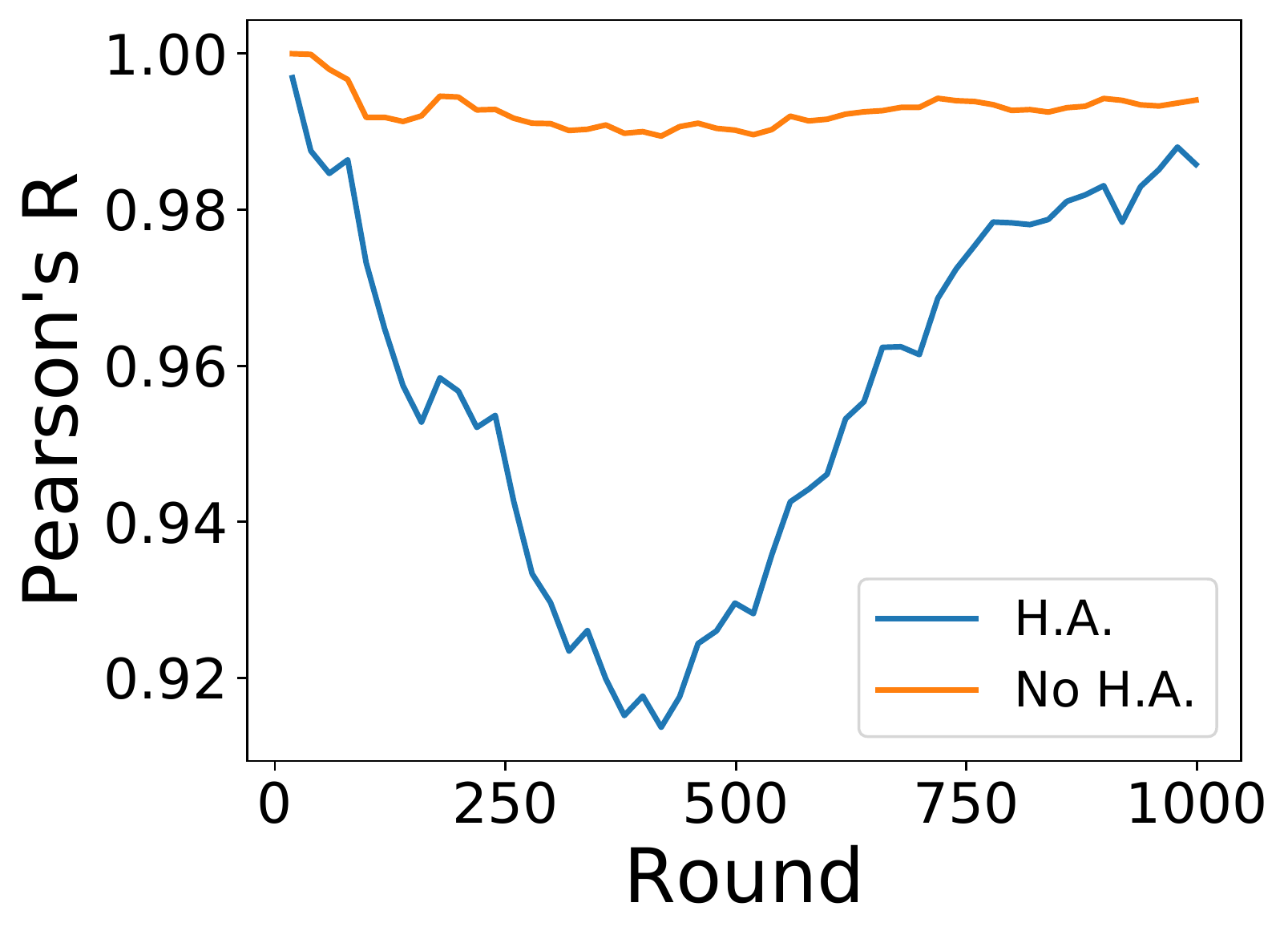}
}%
\subfigure[]{
\label{fig:fil:cnn1_pearson}
\includegraphics[width=0.568\columnwidth]{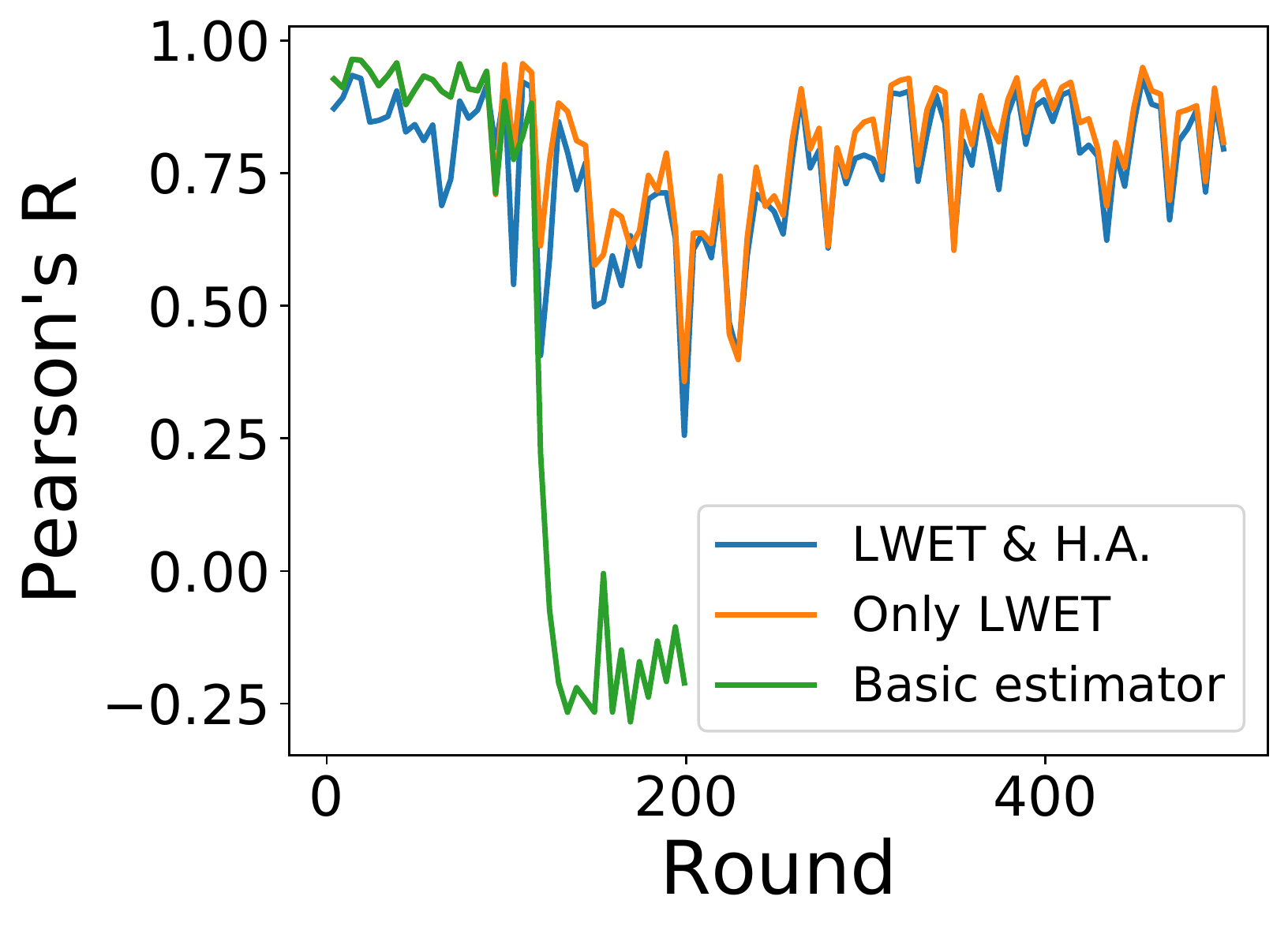}
}%
\subfigure[]{
\label{fig:fil:cnn2_pearson}
\includegraphics[width=0.568\columnwidth]{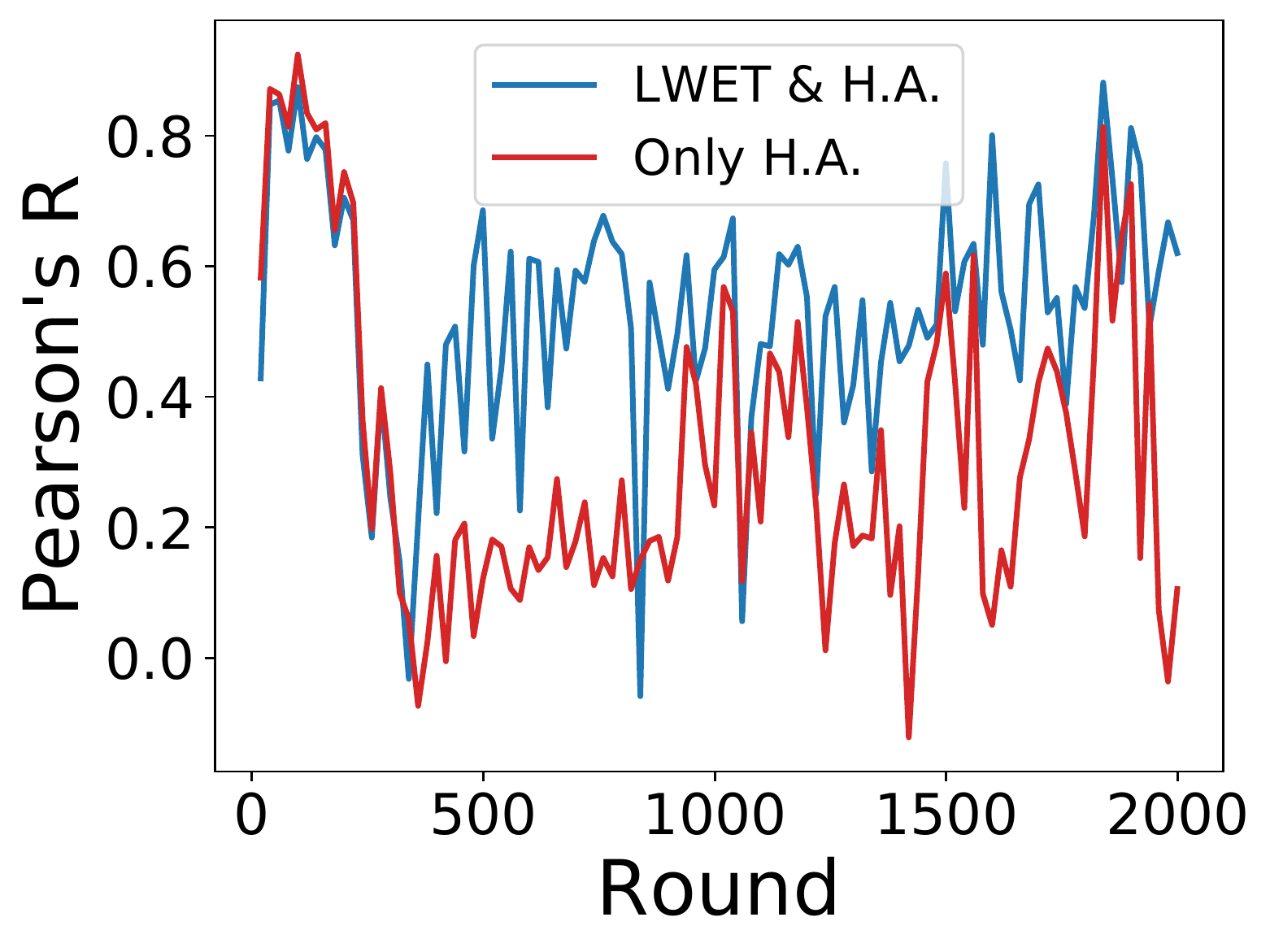}
}%

\subfigure[]{
\label{fig:fil:logreg_scatter}
\includegraphics[width=0.568\columnwidth]{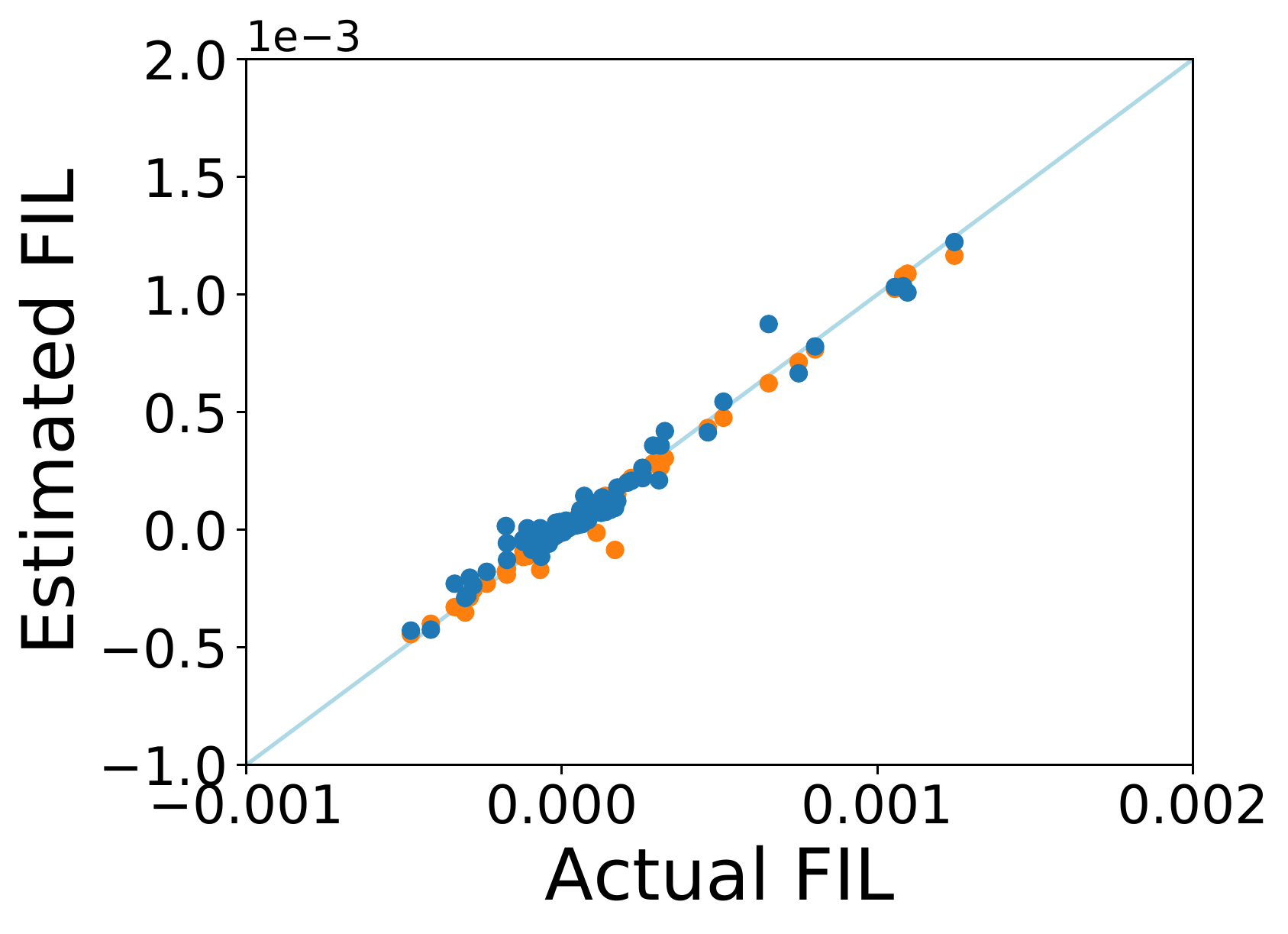}
}%
\subfigure[]{
\label{fig:fil:cnn1_scatter}
\includegraphics[width=0.568\columnwidth]{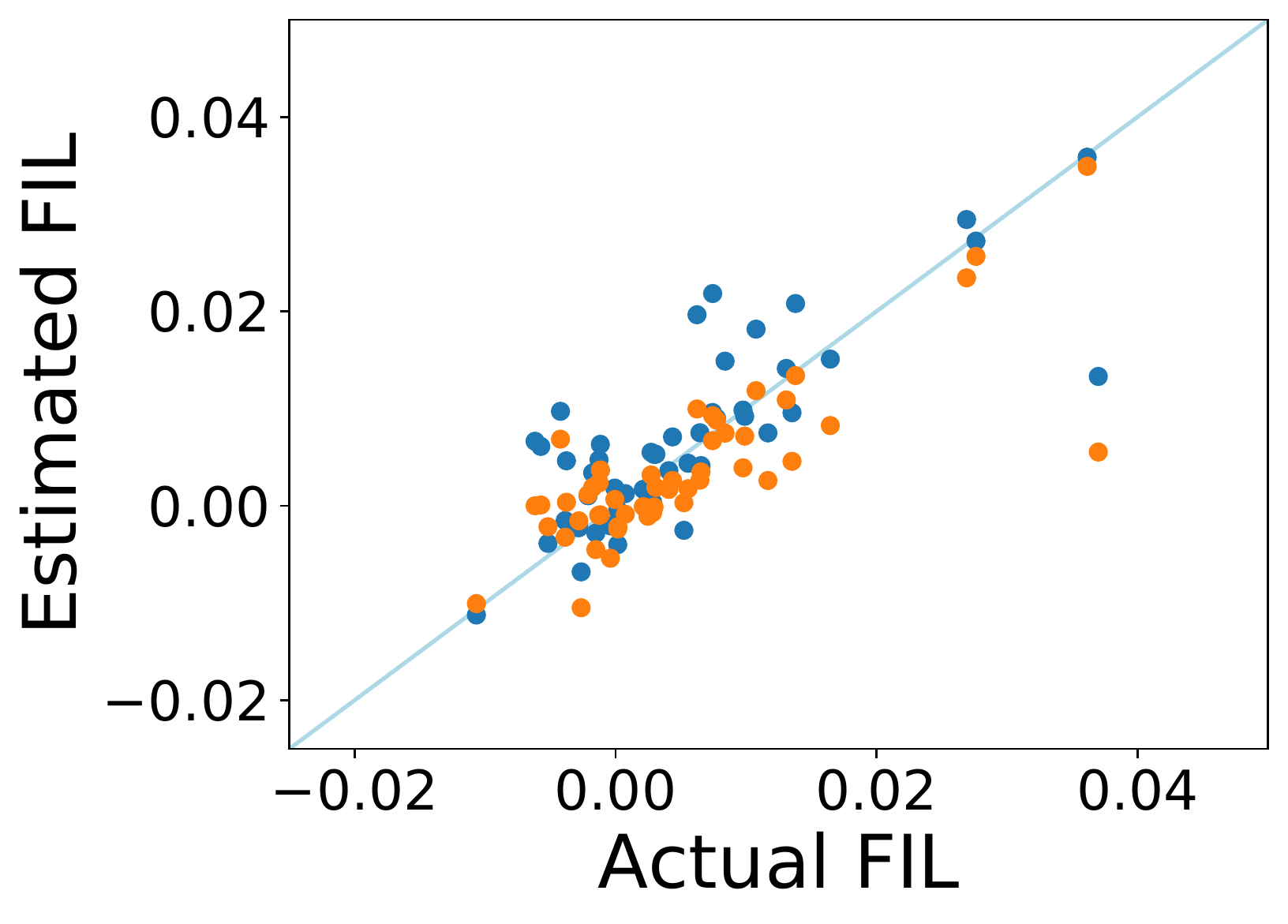}
}%
\subfigure[]{
\label{fig:fil:cnn2_scatter}
\includegraphics[width=0.568\columnwidth]{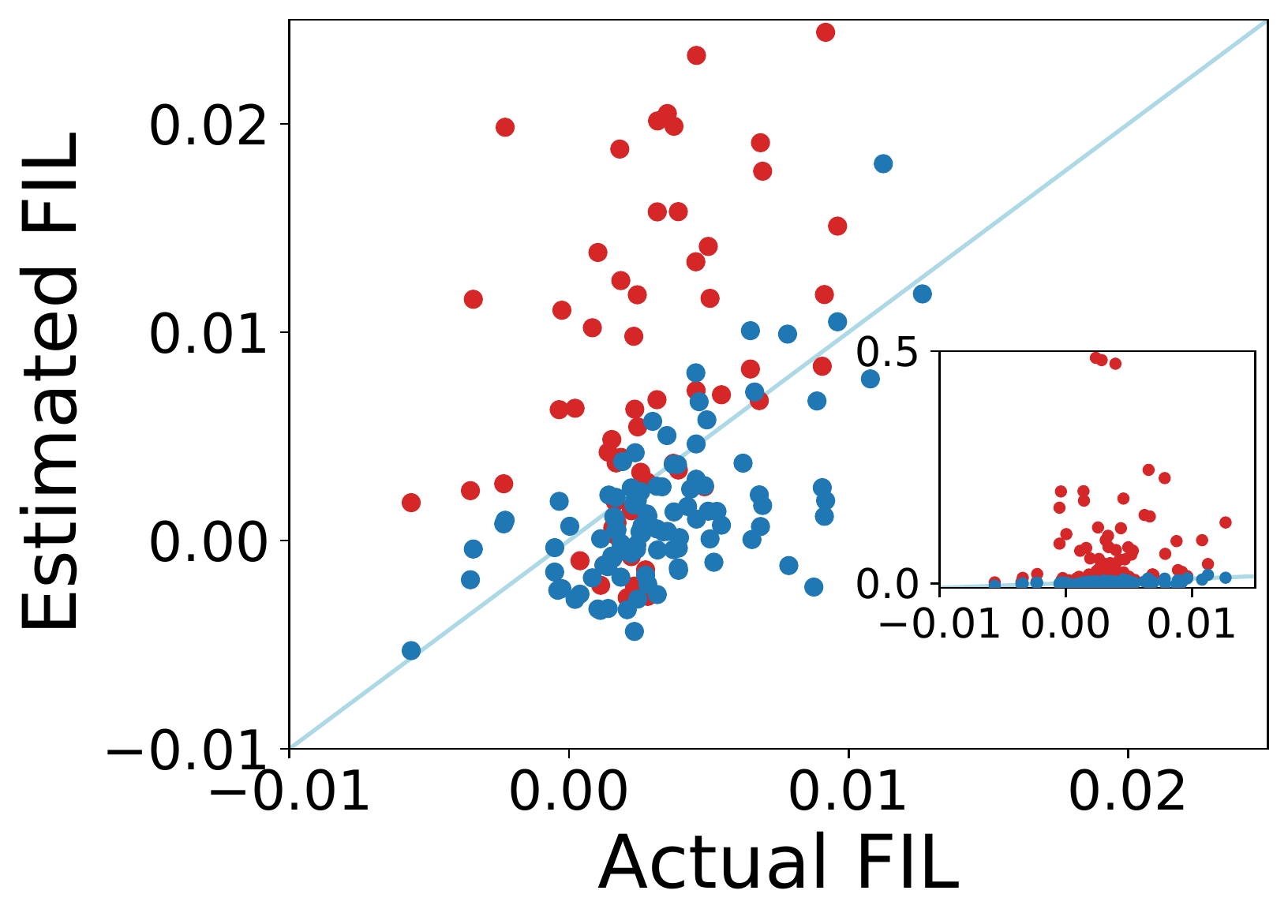}
}%
\centering
\caption{(a)(b)(c) The change of Pearson coefficient over rounds in setting 1, 2 and 3. (d)(e)(f) Estimated FIL vs. actual FIL in settings 1, 2 and 3. The inset in (f) shows the results with a wider y-axis range. In setting 1, we counted 200 clients that were randomly selected from the 1000 clients. In settings 2 and 3, we counted all the clients (50 and 100 clients, respectively).}
\label{fig:fil}
\end{figure*}

In this section, we give a more intuitive demonstration of the experimental results by mapping the high-dimensional FIP to a scalar, FIL. The exact FIL is obtained from the result of leave-one-out test.

By adding estimated FIP to the original global model, we get a estimation of the model trained with one client removed ${\bf w}_t+\epsilon_t^{-c}$. Then we test it on $\mathcal{D}_{test}$ and subtract from the result the loss of the original model to get the estimated FIL, {\em i.e.}, $\mathcal{L}({\bf w}_t+\epsilon_t^{-c}, \mathcal{D}_{test})-\mathcal{L}({\bf w}_t, \mathcal{D}_{test})$.

We compare the estimated FIL with the exact FIL and show the results in Fig.~\ref{fig:fil}. The correlation between the estimated and exact FIL is measured by Pearson's correlation coefficient, and we plot its variation with time in Figs~\ref{fig:fil:logreg_pearson}, \ref{fig:fil:cnn1_pearson}, and \ref{fig:fil:cnn2_pearson}. We also visualize the correlation in the last round in Figs~\ref{fig:fil:logreg_scatter}, \ref{fig:fil:cnn1_scatter}, and \ref{fig:fil:cnn2_scatter}. In particular, the proposed method achieves a Pearson correlation coefficient of 0.6200 at the last round in setting 3, the most difficult setting; in setting 1 Pearson correlation coefficient is 0.9857 and in setting 2 it is 0.7957.

\begin{figure}[htbp]
\centering
\subfigure[]{
\label{fig:eval_and_clean:loss_fil}
\begin{minipage}[t]{0.48\columnwidth}
\centering
\includegraphics[width=1\columnwidth]{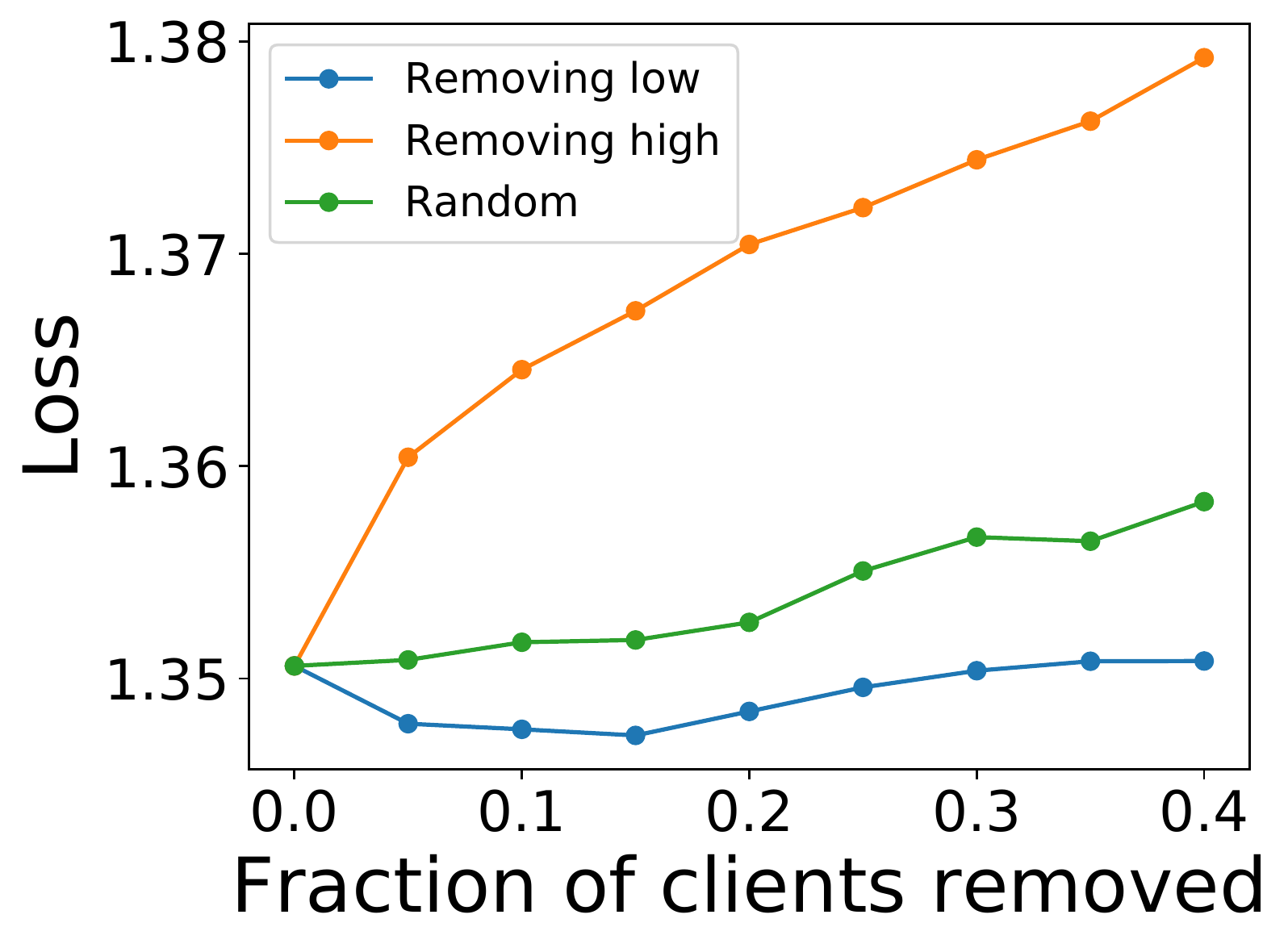}
\includegraphics[width=1\columnwidth]{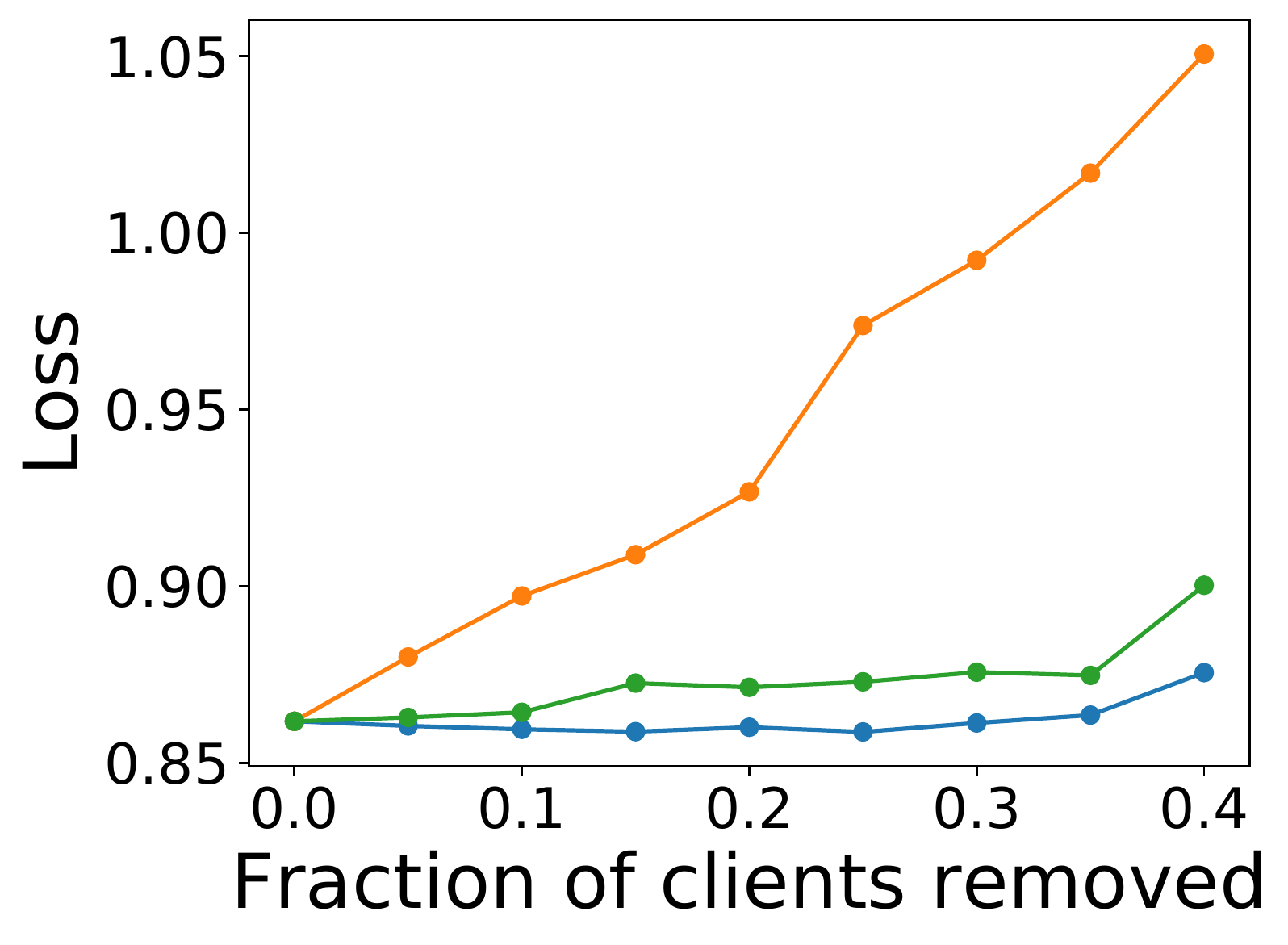}
\end{minipage}
}%
\subfigure[]{
\label{fig:eval_and_clean:acc_fia}
\begin{minipage}[t]{0.48\columnwidth}
\centering
\includegraphics[width=1\columnwidth]{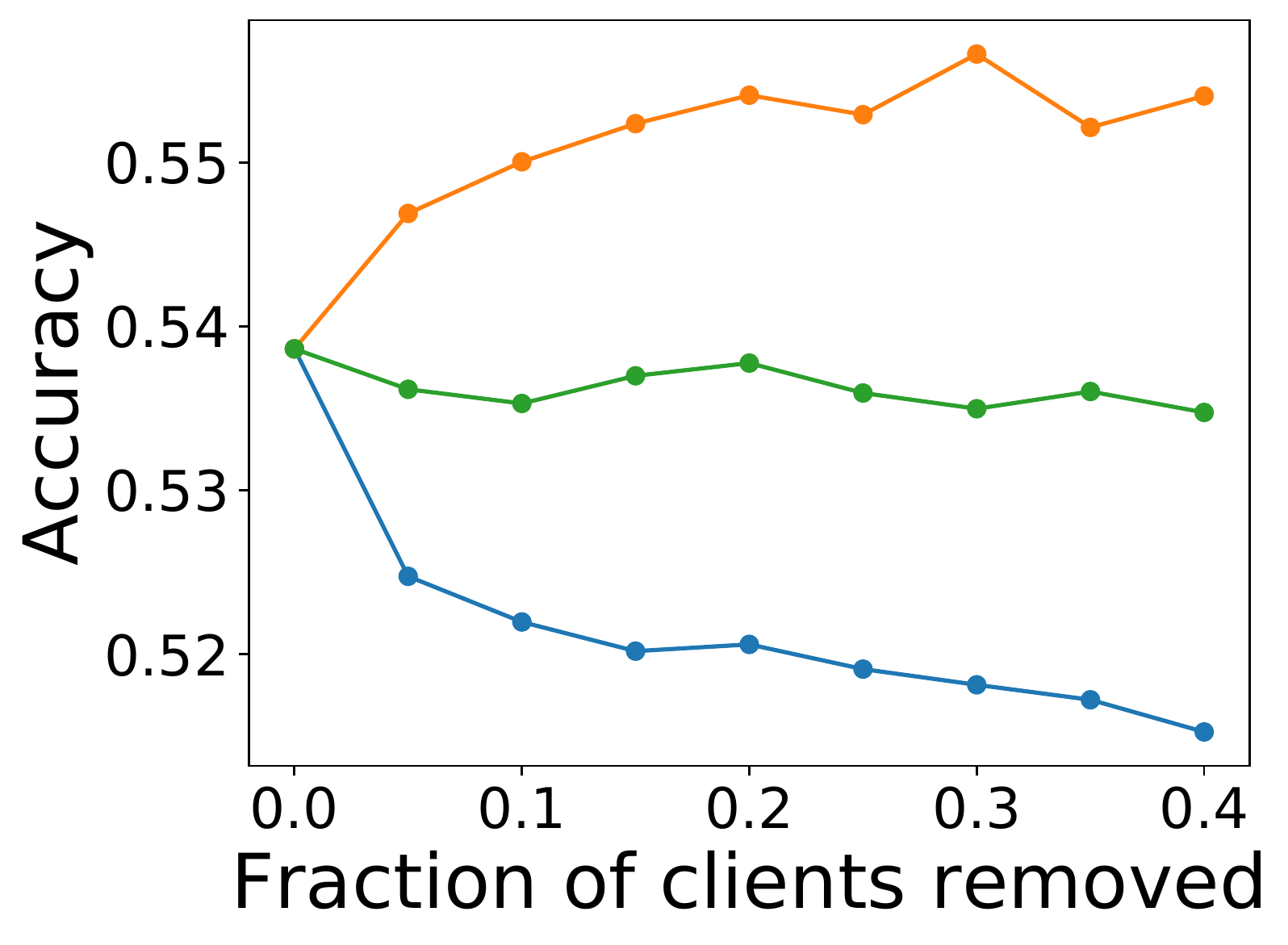}
\includegraphics[width=1\columnwidth]{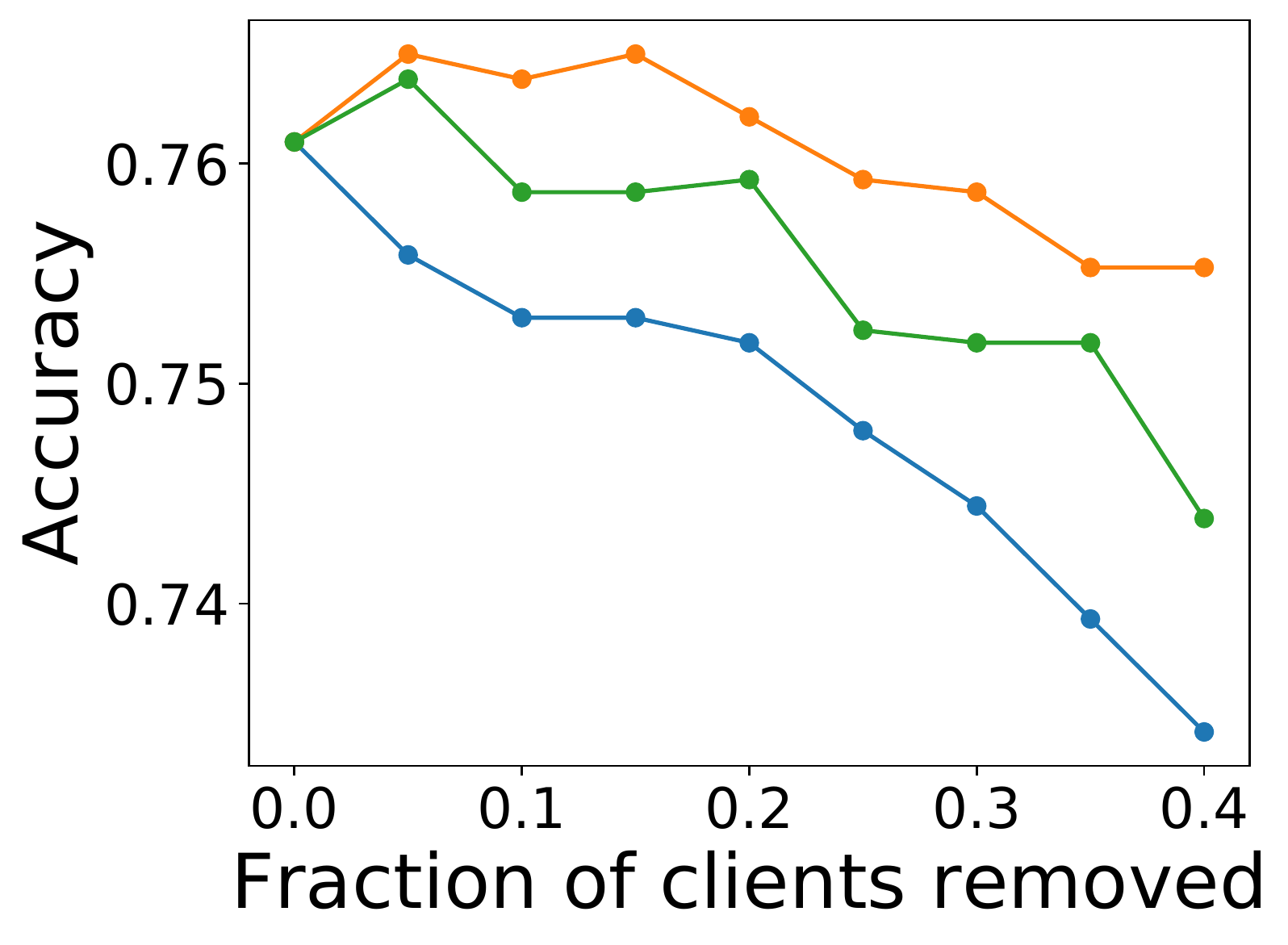}
\end{minipage}
}%
\centering
\caption{We remove clients from the client set at a certain round $x$ in three different ways: from those with lowest influence, from those with highest influence, randomly. (a) We use influence on loss and record the corresponding change in loss. (b) We use influence on accuracy and record the corresponding change in accuracy. Top: The results in setting 1 with $x=700$. Bottom: The results in setting 3 with $x=1500$.}
\label{fig:eval_and_clean}
\end{figure}

\section{Extended Experiments}\label{sec:extension}
In this section, we use FIL and FIA for client valuation and client cleansing, despite that we believe there are more potential applications based on other types of influence given by FIP, for example, the client-level influence on the prediction results.

Influence on model performance can be used as a reasonable metric to determine a client's value. We conduct the following experiment: we make an observation at clients' influence at a certain round, remove some of the clients with highest/lowest influence, from the training set, and then continue the learning process. The experiment is repeated with different fractions of clients removed and the performance of the final global model is recorded each time. As can be seen from Fig.~\ref{fig:eval_and_clean}, removing valuable clients (those with high FIL or low FIA) greatly degrades the model performance, which indicates the importance of these clients. In contrast, removing least valuable clients (those with low FIL or high FIA) improves the model performance. And the curve of randomly removing falls between the other two curves.

The results of removing least valuable clients elicits an application, which we call client cleansing. It is a little different from the data cleansing in \citet{NIPS2019_8674}. Data cleansing is conducted by retraining the model with a subset of data removed. In the setting of federated learning, as we mentioned before, it does not make sense to retrain the model in most real-world scenarios. Therefore, we conduct client cleansing by removing a subset of clients at a certain round and then continue the training. By properly selecting the clients to be removed we can effectively improve the final model. In setting 1, the accuracy is increased from $53.86\%$ to $55.66\%$ by removing $30\%$ of the clients from the client set at round 700. In setting 3, the accuracy is increased from $76.10\%$ to $76.50\%$ by removing $5\%$ of the clients from the client set at round 1500.

\section{Conclusion}
In this paper, we proposed Fed-Influence as a new metric of clients' influence on the global model in federated learning, inspired by the classical statistics notion of influence in centralized learning. The differences between the centralized learning and federated learning create challenges in computing this new metric. To develop an efficient implementation, we first proposed a basic estimator of individual client's influence on parameter, then further revised the estimator and established an algorithm with both robustness and linear cost. It is worth noting that the proposed method is accurate without assumption on convexity or data identity, which is validated by the empirical results on different settings. We also demonstrated how Fed-Influence helps evaluate clients and improve the model performance through client cleansing. Our work provides a quantitative insight into the relationship between individual clients and the global model, we envision which further enhancing the accountability of federated learning.

\section{Acknowledgements}
This work was supported in part by National Key R\&D Program of China No. 2019YFB2102200, in part by China NSF grant No. 62025204, 61972252, 61972254, 61672348, and 61672353, in part by Joint Scientific Research Foundation of the State Education Ministry No. 6141A02033702, and in part by Alibaba Group through Alibaba Innovation Research Program. The opinions, findings, conclusions, and recommendations expressed in this paper are those of the authors and do not necessarily reflect the views of the funding agencies or the government. Z. Zheng is the corresponding author.

\bibliography{ref}

\begin{thebibliography}{24}
\providecommand{\natexlab}[1]{#1}
\providecommand{\url}[1]{\texttt{#1}}
\providecommand{\urlprefix}{URL }
\expandafter\ifx\csname urlstyle\endcsname\relax
  \providecommand{\doi}[1]{doi:\discretionary{}{}{}#1}\else
  \providecommand{\doi}{doi:\discretionary{}{}{}\begingroup
  \urlstyle{rm}\Url}\fi

\bibitem[{Caldas et~al.(2018)Caldas, Wu, Li, Konecn{\'{y}}, McMahan, Smith, and
  Talwalkar}]{DBLP:journals/corr/abs-1812-01097}
Caldas, S.; Wu, P.; Li, T.; Konecn{\'{y}}, J.; McMahan, H.~B.; Smith, V.; and
  Talwalkar, A. 2018.
\newblock {LEAF:} {A} Benchmark for Federated Settings.
\newblock \emph{arXiv preprint arXiv:1812.01097} .

\bibitem[{Cook and Weisberg(1980)}]{8666943c14594ea88ca7f492887941c7}
Cook, R.; and Weisberg, S. 1980.
\newblock Characterizations of an empirical influence function for detecting
  influential cases in regression.
\newblock \emph{Technometrics} 495--508.

\bibitem[{Cook(1977)}]{cook1977detection}
Cook, R.~D. 1977.
\newblock Detection of influential observation in linear regression.
\newblock \emph{Technometrics} 15--18.

\bibitem[{Cook and Weisberg(1982)}]{cook1982residuals}
Cook, R.~D.; and Weisberg, S. 1982.
\newblock \emph{Residuals and influence in regression}.
\newblock New York: Chapman and Hall.

\bibitem[{Friedman, Hastie, and Tibshirani(2001)}]{friedman2001elements}
Friedman, J.; Hastie, T.; and Tibshirani, R. 2001.
\newblock \emph{The elements of statistical learning}.
\newblock Springer series in statistics New York.

\bibitem[{Haddadpour et~al.(2019)Haddadpour, Kamani, Mahdavi, and
  Cadambe}]{haddadpour2019local}
Haddadpour, F.; Kamani, M.~M.; Mahdavi, M.; and Cadambe, V.~R. 2019.
\newblock Local {SGD} with Periodic Averaging: Tighter Analysis and Adaptive
  Synchronization.
\newblock In \emph{Proc. of NeurIPS}, 11080--11092.

\bibitem[{Hamer, Mohri, and Suresh(2020)}]{49366}
Hamer, J.; Mohri, M.; and Suresh, A.~T. 2020.
\newblock {F}ed{B}oost: A Communication-Efficient Algorithm for Federated
  Learning.
\newblock In \emph{Proc. of ICML}, 3973--3983.

\bibitem[{Hampel(1974)}]{hampel1974influence}
Hampel, F.~R. 1974.
\newblock The influence curve and its role in robust estimation.
\newblock \emph{Journal of the american statistical association} 383--393.

\bibitem[{Hara, Nitanda, and Maehara(2019)}]{NIPS2019_8674}
Hara, S.; Nitanda, A.; and Maehara, T. 2019.
\newblock Data Cleansing for Models Trained with {SGD}.
\newblock In \emph{Proc. of NeurIPS}, 4215--4224.

\bibitem[{Jaeckel(1972)}]{jaeckel1972infinitesimal}
Jaeckel, L. 1972.
\newblock \emph{The infinitesimal jackknife}.
\newblock Unpublished memorandum, Bell Telephone Laboratories.

\bibitem[{Kairouz et~al.(2019)Kairouz, McMahan, Avent, Bellet, Bennis, Bhagoji,
  Bonawitz, Charles, Cormode, Cummings et~al.}]{kairouz2019advances}
Kairouz, P.; McMahan, H.~B.; Avent, B.; Bellet, A.; Bennis, M.; Bhagoji, A.~N.;
  Bonawitz, K.; Charles, Z.; Cormode, G.; Cummings, R.; et~al. 2019.
\newblock Advances and open problems in federated learning.
\newblock \emph{arXiv preprint arXiv:1912.04977} .

\bibitem[{Karimireddy et~al.(2019)Karimireddy, Kale, Mohri, Reddi, Stich, and
  Suresh}]{DBLP:journals/corr/abs-1910-06378}
Karimireddy, S.~P.; Kale, S.; Mohri, M.; Reddi, S.~J.; Stich, S.~U.; and
  Suresh, A.~T. 2019.
\newblock Scaffold: Stochastic controlled averaging for on-device federated
  learning.
\newblock \emph{arXiv preprint arXiv:1910.06378} .

\bibitem[{Khanna et~al.(2019)Khanna, Kim, Ghosh, and
  Koyejo}]{khanna2019interpreting}
Khanna, R.; Kim, B.; Ghosh, J.; and Koyejo, S. 2019.
\newblock Interpreting Black Box Predictions using Fisher Kernels.
\newblock In \emph{Proc. of AISTATS}, 3382--3390.

\bibitem[{Koh et~al.(2019)Koh, Ang, Teo, and Liang}]{NIPS2019_8767}
Koh, P.~W.; Ang, K.; Teo, H. H.~K.; and Liang, P. 2019.
\newblock On the Accuracy of Influence Functions for Measuring Group Effects.
\newblock In \emph{Proc. of NeurIPS}, 5255--5265.

\bibitem[{Koh and Liang(2017)}]{DBLP:conf/icml/KohL17}
Koh, P.~W.; and Liang, P. 2017.
\newblock Understanding Black-box Predictions via Influence Functions.
\newblock In \emph{Proc. of ICML}, 1885--1894.

\bibitem[{Li et~al.(2020{\natexlab{a}})Li, Sahu, Zaheer, Sanjabi, Talwalkar,
  and Smith}]{DBLP:conf/mlsys/LiSZSTS20}
Li, T.; Sahu, A.~K.; Zaheer, M.; Sanjabi, M.; Talwalkar, A.; and Smith, V.
  2020{\natexlab{a}}.
\newblock Federated Optimization in Heterogeneous Networks.
\newblock In \emph{Proc. of MLSys}.

\bibitem[{Li et~al.(2020{\natexlab{b}})Li, Huang, Yang, Wang, and
  Zhang}]{DBLP:conf/iclr/LiHYWZ20}
Li, X.; Huang, K.; Yang, W.; Wang, S.; and Zhang, Z. 2020{\natexlab{b}}.
\newblock On the Convergence of FedAvg on Non-IID Data.
\newblock In \emph{Proc. of ICLR}.

\bibitem[{Ly et~al.(2017)Ly, Marsman, Verhagen, Grasman, and
  Wagenmakers}]{ly2017tutorial}
Ly, A.; Marsman, M.; Verhagen, J.; Grasman, R.~P.; and Wagenmakers, E.-J. 2017.
\newblock A tutorial on Fisher information.
\newblock \emph{Journal of Mathematical Psychology} 40--55.

\bibitem[{McMahan et~al.(2017)McMahan, Moore, Ramage, Hampson, and
  y~Arcas}]{DBLP:conf/aistats/McMahanMRHA17}
McMahan, B.; Moore, E.; Ramage, D.; Hampson, S.; and y~Arcas, B.~A. 2017.
\newblock Communication-Efficient Learning of Deep Networks from Decentralized
  Data.
\newblock In \emph{Proc. of AISTATS}, 1273--1282.

\bibitem[{Rothchild et~al.(2020)Rothchild, Panda, Ullah, Ivkin, Stoica,
  Braverman, Gonzalez, and Arora}]{DBLP:journals/corr/abs-2007-07682}
Rothchild, D.; Panda, A.; Ullah, E.; Ivkin, N.; Stoica, I.; Braverman, V.;
  Gonzalez, J.; and Arora, R. 2020.
\newblock Fetchsgd: Communication-efficient federated learning with sketching.
\newblock In \emph{Proc. of ICML}, 8253--8265.

\bibitem[{Wang et~al.(2020)Wang, Liu, Liang, Joshi, and
  Poor}]{DBLP:journals/corr/abs-2007-07481}
Wang, J.; Liu, Q.; Liang, H.; Joshi, G.; and Poor, H.~V. 2020.
\newblock Tackling the objective inconsistency problem in heterogeneous
  federated optimization.
\newblock \emph{Proc. of NeurIPS} .

\bibitem[{Yan et~al.(2020)Yan, Niu, Ding, Zheng, Wu, Chen, Tang, and
  Wu}]{yan2020distributed}
Yan, Y.; Niu, C.; Ding, Y.; Zheng, Z.; Wu, F.; Chen, G.; Tang, S.; and Wu, Z.
  2020.
\newblock Distributed Non-Convex Optimization with Sublinear Speedup under
  Intermittent Client Availability.
\newblock \emph{arXiv preprint arXiv:2002.07399} .

\bibitem[{Yu, Yang, and Zhu(2019)}]{yu2019parallel}
Yu, H.; Yang, S.; and Zhu, S. 2019.
\newblock Parallel Restarted {SGD} with Faster Convergence and Less
  Communication: Demystifying Why Model Averaging Works for Deep Learning.
\newblock In \emph{Proc. of AAAI}, 5693--5700.

\bibitem[{Zhao et~al.(2018)Zhao, Li, Lai, Suda, Civin, and
  Chandra}]{zhao2018federated}
Zhao, Y.; Li, M.; Lai, L.; Suda, N.; Civin, D.; and Chandra, V. 2018.
\newblock Federated Learning with Non-IID Data.
\newblock \emph{arXiv preprint arXiv:1806.00582} .

\end{thebibliography}

\end{document}